\documentclass{article}
\usepackage[preprint]{neurips_2026}
\usepackage[utf8]{inputenc}
\usepackage[T1]{fontenc}
\usepackage{microtype}
\usepackage{graphicx}
\usepackage[table]{xcolor}
\usepackage{booktabs}
\usepackage{longtable}
\usepackage{multirow}
\usepackage{float}
\usepackage{tcolorbox}
\tcbuselibrary{skins}
\usepackage{titlesec}
\usepackage[labelfont=bf,font=small,labelsep=period]{caption}
\usepackage{hyperref}
\usepackage{url}

\definecolor{DeepIndigo}{RGB}{10,20,100}
\definecolor{ElectricCyan}{RGB}{0,120,150}
\definecolor{caspianbg}{RGB}{248,250,255}
\hypersetup{colorlinks=true,linkcolor=DeepIndigo,citecolor=DeepIndigo,urlcolor=DeepIndigo}
\titleformat{\section}{\normalfont\Large\bfseries\color{DeepIndigo}}{\thesection}{1em}{}
\titleformat{\subsection}{\normalfont\large\bfseries\color{DeepIndigo!80}}{\thesubsection}{1em}{}

\usepackage{amsmath,amsfonts,bm}

\def\eqref#1{equation~\ref{#1}}

\def\1{\bm{1}}

\DeclareMathAlphabet{\mathsfit}{\encodingdefault}{\sfdefault}{m}{sl}
\SetMathAlphabet{\mathsfit}{bold}{\encodingdefault}{\sfdefault}{bx}{n}

\begin{document}

\begin{center}
\begin{minipage}[c]{0.25\linewidth}
\includegraphics[height=0.85cm]{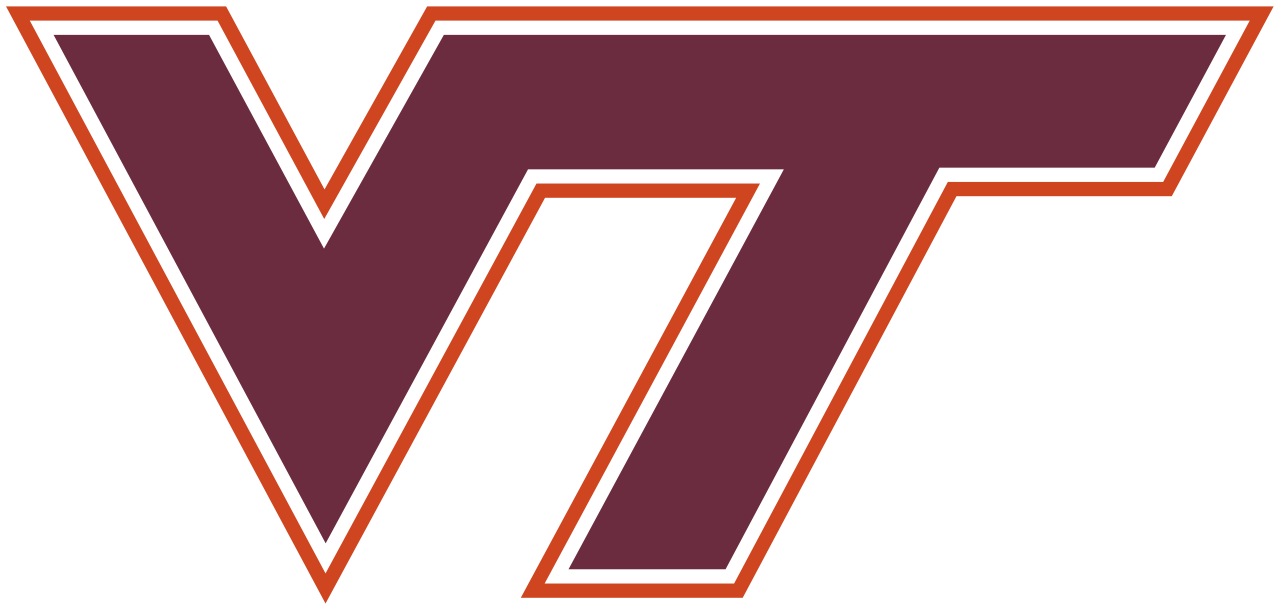}
\end{minipage}%
\begin{minipage}[c]{0.5\linewidth}
\centering
\footnotesize\textcolor{gray!70}{Preprint \\ \today}
\end{minipage}%
\begin{minipage}[c]{0.25\linewidth}
\hfill
\end{minipage}

\vspace{0.25cm}
{\color{DeepIndigo!20}\rule{\linewidth}{0.6pt}}
\vspace{0.4cm}
\end{center}

\begin{center}
{\LARGE\bfseries Evo2Team: When Do Evolved Skills Transfer?\\
From Selection to Deployment\par}
\vspace{0.25cm}
{\large Renxiang Wang$^{1}$ \quad Jiaming Cui$^{1}$\par}
\vspace{0.05cm}
{$^{1}$\texttt{Virginia Tech, Blacksburg, VA}\par}
{\small\texttt{\{renxiangw, jiamingcui\}@vt.edu}\par}
\end{center}

\begin{center}
\begin{tcolorbox}[
  enhanced,
  colback=caspianbg,
  colframe=gray!5,
  arc=0mm,
  outer arc=0mm,
  width=\linewidth,
  left=4mm,
  right=4mm,
  top=3mm,
  bottom=3mm,
  boxrule=0pt,
  borderline west={2.5pt}{0pt}{DeepIndigo},
  before skip=10pt,
  after skip=20pt
]
\textbf{Abstract}\par\smallskip
A skill bank that helps one multi-agent system may leave another's behavior
unchanged. A transferred rule helps only when target agents act on it
successfully. We study this path for
routing and communication skills in Count-Frequency and AgentsNet, using
teams of 4--32 agents and GPT and Qwen model ladders. Source evolution
meets a joint quality, cost, model-tier, and confirmation goal in 14 of
16 settings. We then evaluate Evo2Team, which selects, adapts, and confirms
source skills for the target team, alongside six frozen selectors across
28 transfer directions. Evo2Team's target-side exploration cost is below
that of evolving a new target bank in every direction, even when reused
reference evaluations are charged once. Twenty of 28 held-out outcomes meet the positive-transfer
criterion, including three saved diagnostic tests. Selection alone does not
explain these outcomes: KNN and
CORAL choose different banks in two AgentsNet directions but produce
identical recorded executions. When Evo2Team changes execution, gains can
reach many tasks, as in a Count-Frequency direction that improves 28 of 32
tasks over KNN. Seven positive AgentsNet outcomes save 6.1--14.6\% in
deployment cost while using transferred skills on only three to six of
fifteen tasks. In five earlier accepted directions, all 22 task records
using transferred skills pass three fixed-graph confirmations, but four
fail in recorded executions on new graphs. Graphs and model responses
change together in this comparison. These results show that skill transfer
must be assessed through the actions agents take, the tasks those actions
reach, and the quality and cost of the final deployment.
\end{tcolorbox}
\end{center}

\newcommand{\skillPipelineFigure}{%
\begin{figure}[H]
\centering
\setlength{\abovecaptionskip}{8pt}
\includegraphics[width=\textwidth]{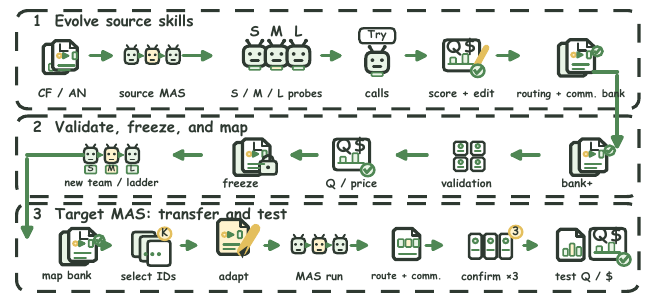}
\caption{\textbf{Evo2Team: from source skill evolution to target deployment.}
A source MAS probes small, medium, and large models, then updates routing
and communication rules using quality and cost feedback. Validation freezes
the bank before it is mapped to a new team. Selection, adaptation, target
execution, three confirmation runs, and held-out testing follow in order.
$Q$ and \$ denote quality and deployment cost.}
\label{fig:skill-pipeline}
\end{figure}%
}

\section{Introduction}
\label{sec:introduction}

Prior work shows how reusable skills guide agents and how a multi-agent
system (MAS) can divide tasks and communicate. Voyager stores learned
procedures in an executable skill library \citep{wang2023voyager}, while
Agent Workflow Memory retains reusable workflows \citep{wang2025awm}.
Multi-agent frameworks organize collaboration through roles and messages
\citep{li2023camel,qian2024chatdev,hong2024metagpt}. Feedback can further
refine instructions and help build reusable skills
\citep{agrawal2026gepa,xia2026skillrl,ma2026skillgen}. However,
transferring a skill bank to a MAS with a different model ladder or team
size does not guarantee that its rules will improve the new team's
decisions. As the team grows, agents may need different communication
patterns, while assigning tasks to smaller or larger models affects both
quality and cost
\citep{qian2025scaling,grotschla2025agentsnet,wang2025agentdropout}.
We therefore study two kinds of skills: routing rules choose a small,
medium, or large model for an agent's task, while communication rules
govern coordination, message budgets, and failure recovery. A target
planner must turn either kind of rule into a concrete action before it
can help (Fig.~\ref{fig:skill-pipeline}). This raises a practical question:
\emph{When we transfer skills, how does the target MAS use them, which
tasks benefit, and do those benefits hold on new instances?}

We study this question in hierarchical Count-Frequency (CF) aggregation
and decentralized AgentsNet coordination, with team sizes from 4 to 32
agents and GPT and Qwen model ladders. Within each task family, we first
develop a skill-evolution procedure and check the resulting bank on
held-out source tasks. The selected policy meets the joint source goal
in 14 of 16 settings. We then transfer the bank to a MAS with a different
team size or model ladder, compare six existing ways to select its skills
under the same target executor, and adapt combinations of those skills.
We call this evolution-to-target procedure Evo2Team.
Recent work learns reusable routines and skills
\citep{wang2025awm,xia2026skillrl} and tests whether these skills transfer across
models \citep{he2026skillcommit}. We examine what happens when such rules
reach a new multi-agent team: which decisions the agents execute, how
many tasks those decisions affect, and whether quality and cost gains
hold on new tasks. We also seek transfers that preserve performance while
costing less per direction to explore than evolving a new target skill bank.
Across all 28 completed directions, Evo2Team's marginal exploration cost is
below fresh target evolution.

\skillPipelineFigure

Our experiments show that useful source skills do not automatically help a new team. We introduce Evo2Team, an end-to-end framework that follows evolved skills from source-team learning through target selection, adaptation, execution, and held-out outcomes. We make three contributions. First, we evaluate Evo2Team's evolved routing and communication skills across team sizes and model ladders in CF and AgentsNet, comparing six selectors under the same executor. Second, we show that Evo2Team costs less to explore than new target evolution in every completed direction, while selecting different skills does not necessarily change what agents do. We also compare feedback-guided KNN and CORAL controls under the same allocated evaluation ceilings. Third, we trace executed skills through task outcomes and new instances, showing where gains reach many tasks, where they save cost on a few, and where fixed-graph confirmation misses failures.

\section{Related Work}
\label{sec:related-work}

Skill evolution builds on a simple idea: keep useful experience and revise
instructions when they fail. Reflexion stores verbal feedback, while
Voyager and Agent Workflow Memory retain routines that can be used again
\citep{shinn2023reflexion,wang2023voyager,wang2025awm}. Self-Refine and ProTeGi use feedback to edit instructions
\citep{madaan2023selfrefine,pryzant2023protegi}. OPRO, EvoPrompt, and
Promptbreeder search over candidate prompts
\citep{yang2024opro,guo2024evoprompt,fernando2024promptbreeder}. DSPy optimizes programs
with multiple model calls, and GEPA uses feedback from execution traces
to improve prompts \citep{khattab2024dspy,agrawal2026gepa}. More recent
work evolves skill banks alongside agent policies, checks whether new
skills help, or verifies when a skill can be applied more broadly
\citep{xia2026skillrl,ma2026skillgen,he2026skillcommit}. These advances
help create the source skills whose transfer we examine.

In a multi-agent system, reusable instructions also decide how agents
work together. CAMEL and AutoGen organize conversations
\citep{li2023camel,wu2024autogen}, while ChatDev and MetaGPT give agents
specialized roles \citep{qian2024chatdev,hong2024metagpt}.
MacNet studies how collaboration changes as teams grow, and AgentsNet
provides coordination tasks across network sizes
\citep{qian2025scaling,grotschla2025agentsnet}. Other work searches
workflows or agent connections \citep{zhang2024aflow,zhuge2024gptswarm}
and reduces communication overhead
\citep{zhang2025agentprune,wang2025agentdropout,chen2025optima}.
Those choices affect both the work agents perform and the cost of
coordinating them, making routing and communication natural skills to
study when the target team changes.

Transfer research asks how knowledge from a source setting can help in a
new one. KMM changes how much each source example counts, and CORAL
aligns the source and target feature distributions
\citep{huang2006correcting,sun2016coral}. Options reuse behaviors with
conditions for starting and stopping them, while successor features reuse
policies when task rewards change
\citep{sutton1999options,barreto2017successor}. FrugalGPT and RouteLLM
address the related choice of which model handles a request
\citep{chen2023frugalgpt,ong2025routellm}. More directly, SkillCommit
studies how skills carry across model families, while TopoPrior transfers
communication priors between domains
\citep{he2026skillcommit,zhang2026topoprior}. Our six selectors draw on
these transfer ideas to choose from the same evolved bank. We also
include a numeric scaling control inspired by DoRA
\citep{liu2024dora}.

The remaining question is what happens after a skill moves to a new
MAS. Source skill evolution can show that a rule helps one team, but it
does not establish that another team can use that rule to improve its
decisions. Even when agents change their behavior, the gain may reach only
a few tasks or disappear on new instances. We therefore follow evolved
routing and communication skills through selection, agent actions,
task-level quality and cost, and new-task outcomes as the team size or
model ladder changes.

\section{Method}
\label{sec:method}

Evo2Team follows the path in Fig.~\ref{fig:skill-pipeline} from evolving skills on
a source MAS to using them in a new team. On the source team, agents try
different model tiers, use quality and cost feedback to revise routing
and communication rules, and validate the resulting bank before freezing
it. We then map that bank to a different team size or model ladder, select
and adapt skills, and let the target agents execute them. Finally, we
confirm the chosen policy three times and measure quality and cost on new
tasks. The figure's numbered rows show this workflow. Within its target
step, Stages I--III ask whether selected skills change agent actions,
whether those actions help enough tasks, and whether the benefit holds
beyond the confirmation tasks.

\subsection{Source skill evolution and validation}
\label{sec:source-evolution}

Source agents try small, medium, and large models under the same upstream
context. Their trial calls reveal how model choice affects quality and
cost, so we use that evidence to revise the routing rules. Task outcomes
also guide changes to coordination, message limits, and recovery rules.
After these updates, separate source tasks are used to choose and confirm
a deployed policy against all-large execution. We freeze the resulting
skill bank and source policy before mapping them to the target team, so
all transfer selectors start from the same source skills.

\subsection{Mapping the bank to the target team}
\label{sec:transfer-objects}

A deployment setting specifies an agent count $n$ and a model ladder
$\mathcal{M}$, which assigns concrete models to small, medium, and large
tiers. Transfer stays within each task family: a routing rule keeps its
relative model tier when mapped to the new ladder, while conditions that
depend on the number of agents are adjusted for the new team. Communication
skills retain their coordination actions, message limits, and other
parameters. Both kinds of skills retain their source evidence. We call
the chosen set of skill IDs $I$, together with any numeric parameter
changes $\theta$, a \emph{recipe} $r=(I,\theta)$. For task $x$, the target
planner first produces a plan and the agents then execute it:
\begin{equation}
 r=(I,\theta)\xrightarrow{P_t(\cdot,x)}p_r(x)
 \xrightarrow{E_t(\cdot,x)}(\tau_r(x),y_r(x)).
 \label{eq:transfer-chain}
\end{equation}
Here $p_r$ is the initial plan, $\tau_r$ is the recorded execution, and
$y_r$ is the output. We use adaptation tasks $\mathcal{D}_a$ to build
recipes, separate confirmation tasks $\mathcal{D}_c$ to choose among them,
and test tasks $\mathcal{D}_t$ to evaluate the final choice. Target model
weights remain fixed throughout.

\subsection{Stage I: The selection-to-execution boundary}
\label{sec:recipe-search}

After mapping the bank, each frozen selector ranks eligible source skills
and includes the top $K$ in the target bank. Rank decides which skills are
available, but the planner reads them in source-bank order and executes a
rule only when its condition applies. If the bank has at most $K$
eligible skills, every selector includes all of them. Even when selections
differ, the planner may still make the same decisions. We therefore compare
which models agents were assigned, which communication settings took
effect, and the full node traces on the same tasks. Full traces also capture
decisions made after agents start responding, with method labels excluded
and numeric formatting normalized for the comparison.

\subsection{Stage II: The execution-to-benefit boundary}
\label{sec:composition}

As in Fig.~\ref{fig:skill-pipeline}, selection supplies an initial bank,
and adaptation tries new combinations before each target run. Starting
from the mapped source policy and a bank of top-ranked skills, it adds or
removes skills, adjusts routing boundaries and numeric communication
settings, and recombines promising recipes. These are discrete edits: skills
can be added, removed, or replaced, routing boundaries move to midpoints
between observed feature values, and integer communication settings are
adjusted. Proposals with the same initial plan as an
already evaluated recipe are skipped before model calls, while the rest are
screened on a subset of $\mathcal{D}_a$ and then evaluated more fully.
The agents' actual task outcomes decide which recipes proceed
(Appendix~\ref{app:implementation}).

Adaptation may also use a different recipe for each group of tasks. CF
groups tasks by difficulty, while AgentsNet groups them by task type and
graph structure. One recipe runs for an entire task so that its agent
interactions stay together. A group whose quality falls below the allowed
level uses all-large instead. We send both a single recipe for every task
and these group-specific choices to confirmation
(Appendix Fig.~\ref{fig:context-guard-detail}).

The number of tasks that use transferred skills matters as much as the
gain within those tasks. Let $T$ be the test tasks running a transferred
recipe and $N$ the total number of test tasks. For task loss or cost $m_i$,
define $\Delta_i=m_i(L)-m_i(r)$ relative to all-large $L$. When the other
tasks use all-large, the average gain is
\begin{equation}
 \overline{\Delta}=\frac{1}{N}\sum_{i\in T}\Delta_i
 =\frac{|T|}{N}\overline{\Delta}_T .
 \label{eq:coverage-gain}
\end{equation}
The second expression applies when at least one task uses a transferred
recipe, and the overall gain is zero when no task does. An improvement on
only a few tasks can therefore have a small overall effect even when the
gain on each is large. We also retain individual task differences because
an average can hide both wins and losses.

\subsection{Stage III: Confirmation and held-out testing}
\label{sec:confirmation}

We compare each recipe with all-large $L$ and with the source policy $F$
executed on the target. The quality gap from a reference $b$ is
\begin{equation}
 g(r,b)=
 \begin{cases}
 R(r)-R(b),&\text{CF},\\
 \max\{S(b)-S(r),P(b)-P(r)\},&\text{AgentsNet},
 \end{cases}
 \label{eq:regression}
\end{equation}
where $R$ is mean task root mean squared error (RMSE) in the CF histogram
counts. For AgentsNet, $S$ is the fraction of tasks solved by the whole
team, while $P$ averages a task-specific score for partly correct outputs.
Lower $R$ and higher $S$ or $P$ indicate better quality. Search first
prioritizes clear quality improvement over $L$, regardless of price, and
then accepts recipes with $g(r,L)\leq\epsilon$ and per-task cost
$c(r)\leq(1+\eta)c(F)$. Other candidates are ranked by how far they miss
these targets. Only $F$'s recorded target price enters this choice, while the quality
of $F$ is checked after the recipe is frozen.

After target agents score candidate recipes, promising ones run three times on separate confirmation tasks $\mathcal{D}_c$. CF combines repeated RMSE values and limits per-run regression; AgentsNet sends task groups that regress to all-large. Confirmation fixes recipes and task-group choices before testing, with full acceptance rules in Appendix~\ref{app:implementation}. Since confirmation tasks stay fixed while test instances change, the same task records can be traced through both stages. A held-out transfer is positive if adaptation costs less than new target evolution and the test outcome is either better quality than both $L$ and $F$, similar quality at the specified price ceiling, or similar quality at lower cost than both. For cost saving, ``similar'' means an RMSE gap at most $\epsilon=0.05$ in CF; in AgentsNet it allows one fewer success among fifteen tasks and a partial-correctness gap at most $\epsilon$. Other completed outcomes are negative. Saved tests of policies not selected by confirmation are diagnostic, not accepted deployments.
\label{sec:measurement}

\section{Results}
\label{sec:results}

\begin{table}[H]
\centering
\footnotesize
\renewcommand{\arraystretch}{0.75}
\setlength{\tabcolsep}{7.5pt}
\caption{Source evolution and transfer across CF and AgentsNet. CF quality
is mean RMSE ($\downarrow$); AgentsNet quality is success/partial correctness
($S/P$, rounded percent, $\uparrow$). Large is all-large, Frozen the mapped
source policy. Policy is source-evolved in Evolve rows and Evo2Team in
Transfer rows. Costs are test USD.
Explore/evolve includes reused references and divides transfer search cost
by new target evolution cost. G/Q denote GPT/Qwen. Y/N mark source goals,
and P/N mark held-out transfer outcomes, including saved diagnostics.}\label{tab:main-results}
\begin{tabular*}{\linewidth}{@{\extracolsep{\fill}}lrrrrrrrc@{}}
\toprule
\multirow{2}{*}{Setting} & \multicolumn{3}{c}{Performance} &
\multicolumn{3}{c}{Test cost (USD)} &
\multirow{2}{*}{\scriptsize\shortstack{Explore/\\evolve}} &
\multirow{2}{*}{Outcome} \\
\cmidrule(lr){2-4}\cmidrule(lr){5-7}
& Large & Frozen & Policy & Large & Frozen & Policy & & \\
\midrule
\multicolumn{9}{@{}l}{\textbf{CF task}\quad\textit{Evolve}} \\
G4 & 0.499 & -- & 0.477 & 0.303 & -- & 0.229 & -- & Y \\
G8 & 0.902 & -- & 0.899 & 0.538 & -- & 0.490 & -- & Y \\
G16 & 2.203 & -- & 1.525 & 1.006 & -- & 0.933 & -- & Y \\
G32 & 4.236 & -- & 4.096 & 1.947 & -- & 1.925 & -- & Y \\
Q4 & 0.667 & -- & 0.568 & 0.227 & -- & 0.171 & -- & Y \\
Q8 & 1.277 & -- & 1.232 & 0.406 & -- & 0.390 & -- & Y \\
Q16 & 3.518 & -- & 1.809 & 0.762 & -- & 0.714 & -- & Y \\
Q32 & 11.951 & -- & 8.987 & 1.481 & -- & 0.295 & -- & Y \\
\addlinespace[1pt]
\cmidrule(lr){1-9}
\multicolumn{9}{@{}l}{\quad\textit{Transfer}} \\
$G4\to G8$ & 0.904 & 1.201 & 0.832 & 0.538 & 0.449 & 1.005 & 0.471 & P \\
$G8\to G16$ & 2.232 & 1.683 & 1.390 & 1.005 & 0.962 & 2.127 & 0.396 & P \\
$G16\to G32$ & 4.217 & 2.960 & 2.161 & 1.947 & 1.852 & 1.577 & 0.231 & P \\
$Q4\to Q8$ & 1.246 & 1.007 & 0.964 & 0.406 & 0.345 & 0.335 & 0.325 & P \\
$Q8\to Q16$ & 3.705 & 3.705 & 2.562 & 0.762 & 0.746 & 1.162 & 0.349 & P \\
$Q16\to Q32$ & 11.452 & 5.570 & 5.344 & 1.481 & 1.381 & 1.381 & 0.587 & P \\
$G4\to Q4$ & 0.646 & 0.638 & 0.630 & 0.227 & 0.157 & 0.400 & 0.414 & P \\
$Q4\to G4$ & 0.567 & 0.554 & 0.543 & 0.302 & 0.259 & 0.259 & 0.396 & P \\
$G8\to Q8$ & 1.282 & 0.980 & 0.874 & 0.406 & 0.328 & 0.694 & 0.325 & P \\
$Q8\to G8$ & 0.857 & 0.857 & 0.756 & 0.538 & 0.521 & 1.305 & 0.512 & P \\
$G16\to Q16$ & 3.476 & 1.858 & 1.741 & 0.762 & 0.634 & 1.012 & 0.267 & P \\
$Q16\to G16$ & 2.096 & 1.383 & 1.341 & 1.005 & 1.088 & 1.069 & 0.247 & P \\
$G32\to Q32$ & 11.713 & 11.713 & 3.440 & 1.481 & 1.463 & 2.080 & 0.988 & P \\
$Q32\to G32$ & 4.321 & 4.861 & 4.747 & 1.990 & 0.715 & 0.711 & 0.117 & N \\
\midrule
\multicolumn{9}{@{}l}{\textbf{AgentsNet task ($S/P$)}\quad\textit{Evolve}} \\
G4 & 80/93 & -- & 80/93 & 0.522 & -- & 0.458 & -- & Y \\
G8 & 47/88 & -- & 47/88 & 2.231 & -- & 2.053 & -- & Y \\
G16 & 53/87 & -- & 47/86 & 8.549 & -- & 8.254 & -- & N \\
G32 & 67/88 & -- & 67/86 & 21.811 & -- & 21.305 & -- & Y \\
Q4 & 87/95 & -- & 93/96 & 0.418 & -- & 0.385 & -- & Y \\
Q8 & 67/93 & -- & 67/93 & 1.458 & -- & 1.417 & -- & Y \\
Q16 & 47/91 & -- & 40/84 & 4.452 & -- & 4.277 & -- & N \\
Q32 & 20/64 & -- & 20/64 & 17.475 & -- & 16.792 & -- & Y \\
\addlinespace[1pt]
\cmidrule(lr){1-9}
\multicolumn{9}{@{}l}{\quad\textit{Transfer}} \\
$G4\to G8$ & 60/89 & 60/89 & 60/89 & 2.249 & 2.249 & 2.249 & 0.761 & N \\
$G8\to G16$ & 73/95 & 73/95 & 73/95 & 5.212 & 5.212 & 5.727 & 0.787 & N \\
$G16\to G32$ & 73/96 & 67/90 & 60/90 & 18.491 & 15.685 & 18.422 & 0.593 & N \\
$Q4\to Q8$ & 67/94 & 67/94 & 67/94 & 1.436 & 1.436 & 1.247 & 0.735 & P \\
$Q8\to Q16$ & 33/85 & 33/85 & 33/85 & 4.610 & 4.610 & 3.936 & 0.814 & P \\
$Q16\to Q32$ & 27/73 & 33/79 & 27/73 & 18.139 & 17.648 & 16.289 & 0.771 & N \\
$G4\to Q4$ & 87/87 & 87/87 & 87/87 & 0.434 & 0.411 & 0.378 & 0.711 & P \\
$Q4\to G4$ & 53/75 & 60/78 & 53/75 & 0.472 & 0.498 & 0.443 & 0.733 & P \\
$G8\to Q8$ & 60/90 & 60/90 & 60/90 & 1.389 & 1.348 & 1.206 & 0.809 & P \\
$Q8\to G8$ & 67/92 & 67/92 & 60/91 & 2.210 & 2.064 & 2.124 & 0.846 & N \\
$G16\to Q16$ & 53/92 & 53/86 & 53/92 & 4.487 & 4.039 & 3.867 & 0.701 & P \\
$Q16\to G16$ & 53/88 & 53/88 & 47/88 & 6.324 & 6.246 & 6.898 & 0.816 & N \\
$G32\to Q32$ & 27/72 & 27/71 & 27/72 & 17.762 & 16.551 & 15.820 & 0.737 & P \\
$Q32\to G32$ & 80/96 & 73/90 & 73/90 & 21.898 & 20.690 & 19.621 & 0.597 & N \\
\bottomrule
\end{tabular*}
\end{table}

\newpage

\paragraph{Experimental setting.}
\label{sec:setup}\label{sec:tasks}\label{sec:baselines}\label{sec:metrics}
We study Count-Frequency (CF) aggregation and AgentsNet graph coordination
\citep{grotschla2025agentsnet} with 4, 8, 16, or 32 agents on GPT and Qwen
model ladders. The 16 source settings lead to 28 completed transfer
directions. CF has 32 test tasks per direction, and AgentsNet has 15 graph
tasks spanning five task types and three topologies. We compare six frozen
skill selectors and Evo2Team with all-large execution and the
mapped source policy on separate adaptation, confirmation, and test splits.
CF quality is mean task RMSE, while AgentsNet quality is whole-team success
($S$) and mean partial correctness ($P$). We account for exploration and
test deployment separately and mark saved tests of unselected policies as
diagnostic. Task construction, model IDs, search budgets, and cost accounting
are detailed in Appendix~\ref{app:implementation}.

\subsection{Overall outcomes: source evolution and transfer}

Table~\ref{tab:main-results} gives the overall picture. Source evolution
meets the joint quality, price, model-tier, and confirmation goal in 14 of
16 settings. Of the 28 held-out Evo2Team outcomes, 20 satisfy the positive
criterion, including three saved diagnostic tests of policies that
confirmation did not select, leaving 17 positive outcomes among the 25
confirmation-selected policies. Evo2Team's exploration also costs less than
evolving a new bank on the target in every direction, even when reused
reference evaluations are charged once. The table separates this search
cost from the quality and price of the deployed policy. These are marginal
transfer costs: source-bank construction is excluded from each direction's
budget and is shared when a bank supports multiple transfers.

All eight CF source policies lower both held-out RMSE and deployment cost
relative to all-large. At GPT/16, RMSE falls from 2.203 to 1.525 as cost
falls from \$1.006 to \$0.933. Six of eight AgentsNet source policies meet
the full goal: Qwen/4 raises whole-network success from 0.867 to 0.933 and
lowers cost from \$0.418 to \$0.385. The two 16-agent exceptions cost less
but each solve one fewer task out of 15. These figures describe deployment
rather than source exploration, and AgentsNet/32 uses generated graphs.
Source-bank sizes and validation details appear in
Appendix~\ref{app:source-evolution}.

Strong source results do not ensure the same target outcome. In CF
G16$\to$G32, all six frozen selectors deploy the same bank and achieve
RMSE 2.854, better than all-large's 4.217. In AgentsNet G8$\to$G16,
five selectors solve only one of 15 tasks and KMM solves none, while
all-large and the mapped source policy each solve eleven
(Appendix Tables~\ref{tab:cf-rmse} and
\ref{tab:agentsnet-S-success-rate}). The smaller figures below examine
where this gap between a selected bank and its deployment arises.

\subsection{Evo2Team costs less to explore than target evolution}

\begin{figure}[H]
\centering
\setlength{\abovecaptionskip}{6pt}
\includegraphics[width=\linewidth]{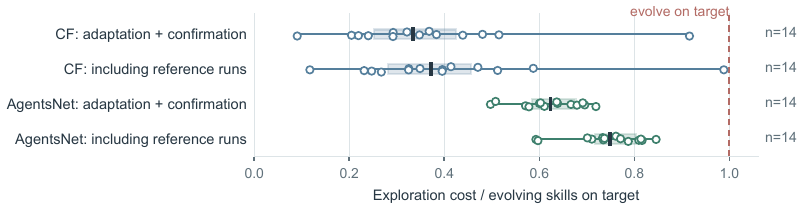}
\caption{\textbf{Evo2Team exploration costs less than target evolution.}
Each circle is one transfer direction (14 per task family). Costs include
adaptation and confirmation, with reused reference evaluations charged once
in the marked rows. Boxes show interquartile ranges, ticks medians, and
lines full ranges. The dashed line marks the cost of new target evolution.}
\label{fig:transfer-cost-outcomes}
\end{figure}

Figure~\ref{fig:transfer-cost-outcomes} shows exploration cost in every
direction. Adaptation and confirmation cost 9.0--91.6\% of
target evolution in CF
and 49.7--71.9\% in AgentsNet. Charging reusable reference runs once still
leaves all ratios below one: 11.7--98.8\% in CF and 59.3--84.6\% in
AgentsNet (Appendix~\ref{app:cost-controls}). Deployment savings then
repay this one-time search cost at different rates. For G4$\to$Q4,
Q4$\to$Q8, and Q4$\to$G4 in AgentsNet, the observed prices require 75,
97, and 209 test-sized batches, respectively, to break even against
all-large. Full comparator results appear in Appendix~\ref{app:full-matrix}.
The six frozen selectors measure selection without target-feedback search.
For a like-budget view of feedback-guided search, the appendix reports KNN
and CORAL controls under the same allocated evaluation ceilings as Evo2Team.
Actual calls and prices are reported separately because deduplication and
confirmation can leave part of a ceiling unused.

\subsection{Different selections can produce the same execution}

A changed set of skills must first alter what agents do. In two AgentsNet
directions, G4$\to$G8 and Q4$\to$Q8, KNN and CORAL select different banks
but produce identical recorded node traces. In all 14 CF directions, they
select the same ordered skills and produce identical traces and task
outcomes. Six CF directions have only 11 eligible skills under a quota of
12, making every selector choose the whole bank. KNN and CORAL agree in the
other eight as well. The AgentsNet cases expose a separate limit: even
different selected banks can lead the planner to the same execution.

\subsection{Executed changes vary in quality, cost, and reach}

\begin{figure}[H]
\centering
\setlength{\abovecaptionskip}{6pt}
\includegraphics[width=\linewidth]{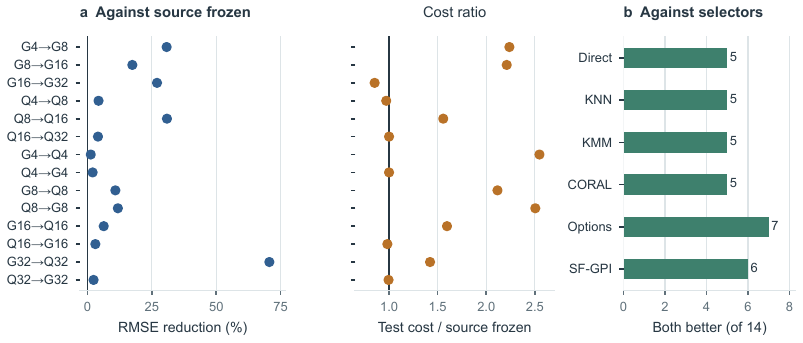}
\caption{\textbf{Evo2Team improves CF quality more consistently than cost.}
(a) Each point represents one direction. Larger RMSE reduction and test-cost
ratios below one favor Evo2Team over the mapped source policy. (b) Counts
of directions where Evo2Team has both lower RMSE and lower deployment
cost than each frozen selector, out of 14. Directions share test tasks.
Exact values are in Appendix
Tables~\ref{tab:cf-rmse} and \ref{tab:cf-evaluation-price-usd}.}
\label{fig:cf-direction-effects}
\end{figure}

Evo2Team lowers CF mean RMSE relative to the mapped source policy in all
14 directions, but lowers deployment cost in only four
(Fig.~\ref{fig:cf-direction-effects}a). Against each frozen selector, it
improves both measures in five to seven directions
(Fig.~\ref{fig:cf-direction-effects}b). Relative to all-large, it lowers
RMSE in 13 directions and cost in five, achieving both in four
(Appendix Tables~\ref{tab:cf-rmse} and
\ref{tab:cf-evaluation-price-usd}). The quality gains are widespread, while
their deployment-price tradeoff depends on the transfer direction.

\raggedbottom

CF G16$\to$G32 shows a broad gain. Relative to all six selectors,
Evo2Team cuts RMSE from 2.854 to 2.161 and improves 28 of 32 tasks over
KNN. It uses four skills instead of eleven, changes a routing threshold,
and moves the team from four to eight groups. Calls fall from 57 to 42 per
task, recorded reroutes from 608 to zero, large-model call share from
89.5\% to 56.9\%, and deployment cost from \$2.763 to \$1.577
(Appendix Fig.~\ref{fig:execution-case}). In Q4$\to$Q8, Evo2Team also
cuts RMSE by 9--11\% and cost by 22--54\% against all six frozen
selectors.

The size of the gain does not reveal how many tasks benefit. G8$\to$Q8
and Q8$\to$G8 each reduce mean RMSE by about 51\% relative to KNN,
winning 31 and 30 of 32 tasks. Q16$\to$Q32, by comparison, reduces
mean RMSE by 4.1\% against KNN but improves only 18 tasks. The saved Q4$\to$G4
diagnostic wins 13 tasks and loses 15. The complete direction-wise
distributions in Appendix Table~\ref{tab:cf-rmse} show why mean quality
and task breadth need to be read together.

Deployment cost adds another distinction. Against the mapped source policy,
G4$\to$G8 lowers CF RMSE from 1.201 to 0.832, but raises the cost of the
32-task test batch from \$0.449 to \$1.005. Q4$\to$Q8 lowers RMSE from
1.007 to 0.964 while reducing that cost from \$0.345 to \$0.335. Both
directions gain quality, but only the latter does so at a lower deployment
price (Appendix Tables~\ref{tab:cf-rmse} and
\ref{tab:cf-evaluation-price-usd}).

\begin{figure}[H]
\centering
\setlength{\abovecaptionskip}{6pt}
\includegraphics[width=\linewidth]{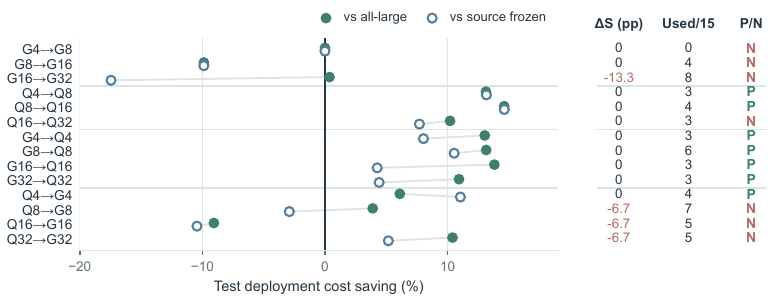}
\caption{\textbf{Evo2Team's AgentsNet deployment outcomes in 14 directions.}
Each row is one transfer direction. Filled and open circles show test-cost
savings against all-large and the mapped source policy. Negative values mean
higher cost. Columns give whole-team success change ($\Delta S$, percentage
points) against all-large, tasks using transferred skills out of 15, and
held-out outcome (P/N). Saved diagnostic tests are included. Exact metrics
are in Appendix~\ref{app:full-matrix}.}
\label{fig:agentsnet-tradeoffs}
\end{figure}

AgentsNet gains concentrate on fewer tasks and often come through cost
(Fig.~\ref{fig:agentsnet-tradeoffs}). Five accepted directions meet the
positive held-out criterion, saving 6.10--14.61\% against all-large while
preserving its quality within the stated tolerance. Two saved diagnostic
directions also meet the criterion. In G4$\to$Q4 and Q4$\to$Q8, the
Evo2Team policy exactly preserves both reference quality scores and saves
13.02\% and 13.13\%, respectively. Only the three consensus tasks in each
direction run transferred skills. Across all ten accepted AgentsNet
directions, transferred recipes run on 45 of 150 direction--task records;
the other 105 use all-large.

Greater coverage is not automatically better. G16$\to$G32 transfers skills
to eight AgentsNet tasks but loses 13.33 percentage points of success while
saving only 0.37\% against all-large. Q16$\to$G16 uses transferred skills
on five tasks, loses 6.67 points of success, and raises cost by 9.06\%.
Coverage identifies where the policy acts; paired quality and cost show
whether those actions help.

\flushbottom
\subsection{Confirmation does not ensure success on new instances}

Confirmation narrows deployment, but it cannot certify every new
execution. In five earlier accepted AgentsNet directions, successive
checks reduce the 36 task records using transferred recipes to 30, 25,
and finally 22 (Appendix Fig.~\ref{fig:confirmation-outcomes}). Each
record is one task in one direction. All 22 retained records succeed in
three runs on fixed confirmation
graphs, whereas 18 succeed in the recorded runs on new test graphs. Three
of the four failures are regressions against both all-large and the mapped
source policy. Both references also fail on the fourth, Q8$\to$G8 coloring
on a Watts--Strogatz graph
(Appendix Table~\ref{tab:guard-trajectory}).

The three regressions occur on Q4$\to$G4 consensus, Q8$\to$G8 coloring,
and Q16$\to$G16 matching (Appendix Table~\ref{tab:branch-failures}).
In Q4$\to$G4, a gain on leader election offsets the consensus loss, so
aggregate success does not change. The failed confirmation--test graph
pairs have the same edge count and round budget but no common change in
maximum degree or diameter. Because the test also draws fresh model
responses, the recorded difference cannot be assigned to graph structure
alone (Appendix Table~\ref{tab:failed-graph-attributes}).

\section{Analysis and Ablations}
\label{sec:analysis}\label{sec:ablations}

The main results separate selecting a skill bank, changing execution, and
improving deployment. Execution records make this distinction concrete: in
CF G16$\to$G32, the mapped source and adapted policies both
use eight groups and 42 calls per task, yet Evo2Team lowers RMSE from
2.960 to 2.161 as the large-model call share falls from 76.2\% to 56.9\%.
In Q16$\to$Q32, calls, groups, model shares, and reroutes match across the
two policies, so these coarse records do not explain its smaller quality
gain. We use ablations to examine which parts of the target procedure
change the observed quality--cost balance.

\subsection{Component ablations}

\begin{figure}[htbp]
\centering
\setlength{\abovecaptionskip}{6pt}
\includegraphics[width=\linewidth]{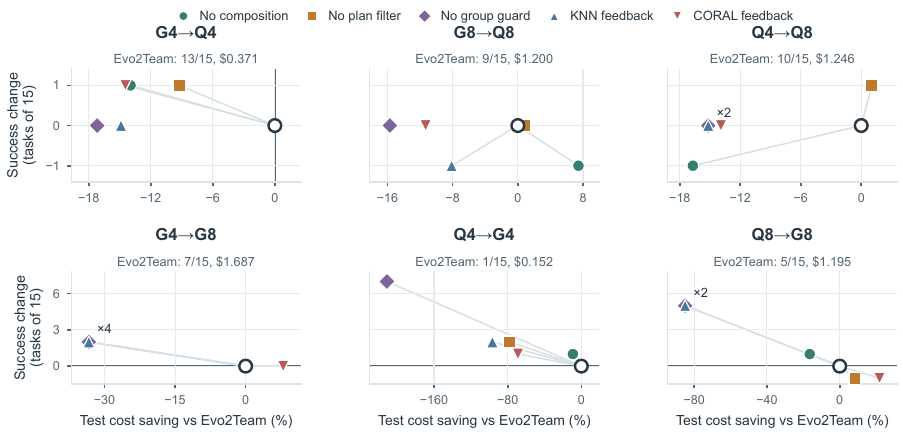}
\caption{\textbf{AgentsNet ablations for Qwen (top) and GPT (bottom) targets.}
Each panel is one transfer direction with 15 test tasks. Points show changes
relative to Evo2Team (open circle): rightward means lower deployment
cost and upward means more whole-team successes. KNN/CORAL feedback share
the allocated search ceiling, though actual API spending can differ.
$\times n$ marks overlapping points. Exact values are in Appendix
Table~\ref{tab:agentsnet-ablation-complete}.}
\label{fig:ablation-agentsnet}
\end{figure}

We compare Evo2Team with component removals and feedback-guided
KNN/CORAL alternatives in three Qwen-target and three GPT-target AgentsNet
directions (Fig.~\ref{fig:ablation-agentsnet}). Every variant runs on the
same 15 graph tasks within its direction, and these runs remain separate
from the main transfer outcomes. Appendix
Table~\ref{tab:agentsnet-ablation-complete} gives success, partial
correctness, and deployment cost for every variant.

The KNN/CORAL feedback controls search skill subsets using target outcomes
under the same allocated feedback ceiling as Evo2Team: 130 adaptation
and 90 confirmation candidate--task evaluations in AgentsNet. Actual API
spending can differ. On all three Qwen targets, Evo2Team costs less
to deploy than either control and matches or exceeds KNN's success. On the
three GPT targets, KNN succeeds on more tasks but costs more to deploy each
time. The comparison therefore exposes a quality--cost tradeoff.

The task-group guard consistently limits spending, but its quality effect
depends on the target. Removing it leaves both quality measures unchanged
in the three Qwen-target directions and raises cost by 15--17\%. For GPT
targets, removing it raises success from 1/15 to 8/15 in Q4$\to$G4 as
cost rises from \$0.152 to \$0.472. In Q8$\to$G8, success rises from
5/15 to 10/15 as cost rises from \$1.195 to \$2.210. The guard therefore
buys lower cost in these six directions, with a target-dependent effect on
success.

The Qwen-target CF ablations show a related quality--cost tradeoff
(Appendix Fig.~\ref{fig:ablation-qwen-cf} and
Table~\ref{tab:cf-ablation-complete}). Removing composition lowers RMSE
in all three directions but increases deployment cost in all three.
Removing the planning filter in Q4$\to$Q8 instead lowers cost from
\$0.720 to \$0.329 as RMSE rises from 0.916 to 0.960.

\section{Discussion and Conclusion}
\label{sec:conclusion}

Moving a skill bank between multi-agent teams is useful when the target
team acts on its rules and those actions help the tasks it encounters.
Source evolution supplies strong starting skills in 14 of 16 settings,
but Evo2Team's results show why the bank's source performance is only
part of the story. Selectors can produce the same execution, improvements
can reach just a few tasks, and three runs on a fixed graph can miss
failures in a recorded execution on a new graph. Evaluating transferred
skills therefore calls for the agents' actual decisions, the number of
tasks those decisions affect, and both task quality and deployment cost.
Evo2Team's marginal exploration cost is below that of evolving a new target
bank in all 28 completed directions, even when reference runs are charged
once. The comparison counts target-side transfer costs, with source-bank
construction shared across directions. Our study covers CF aggregation and
AgentsNet graph coordination, and the next step is to evaluate the same
workflow on longer-horizon tasks. On new graphs, structural changes and fresh
model responses occur together, so the recorded outcomes capture their joint
effect.

\subsection*{AI use statement}

Generative AI tools assisted with exploratory analysis of experiment logs
and interpretation of recorded results, preparation of figures and plotting
code, literature searches, and manuscript organization, drafting,
translation, and language editing. The authors review and verify
AI-assisted analyses, code, figures, citations, and text against the
experimental records and original sources and take responsibility for
the final content of this work.

\subsection*{Ethics statement}

The experiments use synthetic Count-Frequency inputs and graph-coordination
benchmark tasks and do not involve human participants or personal data.
The evaluated skills control model routing and agent communication, so their
deployment effects include changes in answer quality and API cost. We report
both and retain failure cases rather than presenting lower cost alone as a
benefit. Model outputs are used for research evaluation and are not used to
make decisions about people.

\subsection*{Reproducibility statement}

The experimental setup specifies task splits, model ladders, transfer
directions, selector quotas, adaptation budgets, and evaluation metrics.
Appendix~\ref{app:implementation} gives the source-skill mapping, search,
confirmation, acceptance, cache, and cost protocols. Source outcomes,
full direction-wise matrices, and task-record trajectories appear in
Appendix~\ref{app:source-evolution}--\ref{app:confirmation-records}. An
anonymous code archive includes the implementation and analysis scripts
with the compact data used for the figures. Model IDs and execution details
are listed in Appendix~\ref{app:implementation}.

\bibliography{references}

@inproceedings{agrawal2026gepa,
  title = {{GEPA}: Reflective Prompt Evolution Can Outperform Reinforcement Learning},
  author = {Agrawal, Lakshya A and Tan, Shangyin and Soylu, Dilara and Ziems, Noah and Khare, Rishi and Opsahl-Ong, Krista and Singhvi, Arnav and Shandilya, Herumb and Ryan, Michael J and Jiang, Meng and Potts, Christopher and Sen, Koushik and Dimakis, Alexandros G. and Stoica, Ion and Klein, Dan and Zaharia, Matei and Khattab, Omar},
  booktitle = {The Fourteenth International Conference on Learning Representations},
  year = {2026},
  url = {https://proceedings.iclr.cc/paper_files/paper/2026/hash/0e9e708b6f48e14fd0ac29e167413f76-Abstract-Conference.html}
}

@inproceedings{ong2025routellm,
  title = {{RouteLLM}: Learning to Route {LLMs} with Preference Data},
  author = {Ong, Isaac and Almahairi, Amjad and Wu, Vincent and Chiang, Wei-Lin and Wu, Tianhao and Gonzalez, Joseph E. and Kadous, M Waleed and Stoica, Ion},
  booktitle = {The Thirteenth International Conference on Learning Representations},
  year = {2025},
  url = {https://arxiv.org/abs/2406.18665}
}

@inproceedings{qian2025scaling,
  title = {Scaling Large Language Model-Based Multi-Agent Collaboration},
  author = {Qian, Chen and Xie, Zihao and Wang, Yifei and Liu, Wei and Zhu, Kunlun and Xia, Hanchen and Dang, Yufan and Du, Zhuoyun and Chen, Weize and Yang, Cheng and Liu, Zhiyuan and Sun, Maosong},
  booktitle = {The Thirteenth International Conference on Learning Representations},
  year = {2025},
  url = {https://proceedings.iclr.cc/paper_files/paper/2025/file/66a026c0d17040889b50f0dfa650e5e0-Paper-Conference.pdf}
}

@article{wang2023voyager,
  title = {Voyager: An Open-Ended Embodied Agent with Large Language Models},
  author = {Wang, Guanzhi and Xie, Yuqi and Jiang, Yunfan and Mandlekar, Ajay and Xiao, Chaowei and Zhu, Yuke and Fan, Linxi and Anandkumar, Anima},
  journal = {arXiv preprint arXiv:2305.16291},
  year = {2023},
  url = {https://arxiv.org/abs/2305.16291}
}

@article{zhang2026topoprior,
  title = {Learning Transferable Topology Priors for Multi-Agent {LLM} Collaboration Across Domains},
  author = {Zhang, Taolin and Zhou, Zijie and Wan, Jiuheng and Hu, Tingyuan and Wang, Chengyu and He, Xiaofeng and Hong, Richang},
  journal = {arXiv preprint arXiv:2605.17359},
  year = {2026},
  url = {https://arxiv.org/abs/2605.17359}
}

@article{grotschla2025agentsnet,
  title = {{AgentsNet}: Coordination and Collaborative Reasoning in Multi-Agent {LLMs}},
  author = {Gr{\"o}tschla, Florian and M{\"u}ller, Luis and T{\"o}nshoff, Jan and Galkin, Mikhail and Perozzi, Bryan},
  journal = {arXiv preprint arXiv:2507.08616},
  year = {2025},
  url = {https://arxiv.org/abs/2507.08616}
}

@article{shinn2023reflexion,
  title = {Reflexion: Language Agents with Verbal Reinforcement Learning},
  author = {Shinn, Noah and Cassano, Federico and Berman, Edward and Gopinath, Ashwin and Narasimhan, Karthik and Yao, Shunyu},
  journal = {arXiv preprint arXiv:2303.11366},
  year = {2023},
  url = {https://arxiv.org/abs/2303.11366}
}

@article{zhang2024aflow,
  title = {{AFlow}: Automating Agentic Workflow Generation},
  author = {Zhang, Jiayi and Xiang, Jinyu and Yu, Zhaoyang and Teng, Fengwei and Chen, Xionghui and Chen, Jiaqi and Zhuge, Mingchen and Cheng, Xin and Hong, Sirui and Wang, Jinlin and Zheng, Bingnan and Liu, Bang and Luo, Yuyu and Wu, Chenglin},
  journal = {arXiv preprint arXiv:2410.10762},
  year = {2024},
  url = {https://arxiv.org/abs/2410.10762}
}

@inproceedings{huang2006correcting,
  title = {Correcting Sample Selection Bias by Unlabeled Data},
  author = {Huang, Jiayuan and Gretton, Arthur and Borgwardt, Karsten and Sch{\"o}lkopf, Bernhard and Smola, Alex J.},
  booktitle = {Advances in Neural Information Processing Systems},
  volume = {19},
  year = {2006},
  url = {https://proceedings.neurips.cc/paper/2006/hash/a2186aa7c086b46ad4e8bf81e2a3a19b-Abstract.html}
}

@article{sun2016coral,
  title = {Return of Frustratingly Easy Domain Adaptation},
  author = {Sun, Baochen and Feng, Jiashi and Saenko, Kate},
  journal = {Proceedings of the AAAI Conference on Artificial Intelligence},
  volume = {30},
  number = {1},
  year = {2016},
  doi = {10.1609/aaai.v30i1.10306},
  url = {https://ojs.aaai.org/index.php/AAAI/article/view/10306}
}

@article{sutton1999options,
  title = {Between {MDPs} and semi-{MDPs}: A Framework for Temporal Abstraction in Reinforcement Learning},
  author = {Sutton, Richard S. and Precup, Doina and Singh, Satinder},
  journal = {Artificial Intelligence},
  volume = {112},
  number = {1--2},
  pages = {181--211},
  year = {1999},
  doi = {10.1016/S0004-3702(99)00052-1},
  url = {https://www.sciencedirect.com/science/article/pii/S0004370299000521}
}

@inproceedings{barreto2017successor,
  title = {Successor Features for Transfer in Reinforcement Learning},
  author = {Barreto, Andr{\'e} and Dabney, Will and Munos, R{\'e}mi and Hunt, Jonathan J. and Schaul, Tom and van Hasselt, Hado P. and Silver, David},
  booktitle = {Advances in Neural Information Processing Systems},
  volume = {30},
  year = {2017},
  url = {https://proceedings.neurips.cc/paper/2017/hash/350db081a661525235354dd3e19b8c05-Abstract.html}
}

@inproceedings{liu2024dora,
  title = {{DoRA}: Weight-Decomposed Low-Rank Adaptation},
  author = {Liu, Shih-Yang and Wang, Chien-Yi and Yin, Hongxu and Molchanov, Pavlo and Wang, Yu-Chiang Frank and Cheng, Kwang-Ting and Chen, Min-Hung},
  booktitle = {Proceedings of the 41st International Conference on Machine Learning},
  series = {Proceedings of Machine Learning Research},
  volume = {235},
  pages = {32100--32121},
  year = {2024},
  publisher = {PMLR},
  url = {https://proceedings.mlr.press/v235/liu24bn.html}
}

@inproceedings{madaan2023selfrefine,
  title = {Self-Refine: Iterative Refinement with Self-Feedback},
  author = {Madaan, Aman and Tandon, Niket and Gupta, Prakhar and Hallinan, Skyler and Gao, Luyu and Wiegreffe, Sarah and Alon, Uri and Dziri, Nouha and Prabhumoye, Shrimai and Yang, Yiming and Gupta, Shashank and Majumder, Bodhisattwa Prasad and Hermann, Katherine and Welleck, Sean and Yazdanbakhsh, Amir and Clark, Peter},
  booktitle = {Advances in Neural Information Processing Systems},
  volume = {36},
  year = {2023},
  url = {https://proceedings.neurips.cc/paper_files/paper/2023/hash/91edff07232fb1b55a505a9e9f6c0ff3-Abstract-Conference.html}
}

@inproceedings{pryzant2023protegi,
  title = {Automatic Prompt Optimization with ``Gradient Descent'' and Beam Search},
  author = {Pryzant, Reid and Iter, Dan and Li, Jerry and Lee, Yin and Zhu, Chenguang and Zeng, Michael},
  booktitle = {Proceedings of the 2023 Conference on Empirical Methods in Natural Language Processing},
  pages = {7957--7968},
  year = {2023},
  publisher = {Association for Computational Linguistics},
  doi = {10.18653/v1/2023.emnlp-main.494},
  url = {https://aclanthology.org/2023.emnlp-main.494/}
}

@inproceedings{yang2024opro,
  title = {Large Language Models as Optimizers},
  author = {Yang, Chengrun and Wang, Xuezhi and Lu, Yifeng and Liu, Hanxiao and Le, Quoc V and Zhou, Denny and Chen, Xinyun},
  booktitle = {The Twelfth International Conference on Learning Representations},
  year = {2024},
  url = {https://proceedings.iclr.cc/paper_files/paper/2024/hash/3339f19c5fcee3ad74502947a32be9e6-Abstract-Conference.html}
}

@inproceedings{guo2024evoprompt,
  title = {Connecting Large Language Models with Evolutionary Algorithms Yields Powerful Prompt Optimizers},
  author = {Guo, Qingyan and Wang, Rui and Guo, Junliang and Li, Bei and Song, Kaitao and Tan, Xu and Liu, Guoqing and Bian, Jiang and Yang, Yujiu},
  booktitle = {The Twelfth International Conference on Learning Representations},
  year = {2024},
  url = {https://proceedings.iclr.cc/paper_files/paper/2024/hash/9156b0f6dfa9bbd18c79cc459ef5d61c-Abstract-Conference.html}
}

@inproceedings{fernando2024promptbreeder,
  title = {Promptbreeder: Self-Referential Self-Improvement via Prompt Evolution},
  author = {Fernando, Chrisantha and Banarse, Dylan Sunil and Michalewski, Henryk and Osindero, Simon and Rockt{\"a}schel, Tim},
  booktitle = {Proceedings of the 41st International Conference on Machine Learning},
  series = {Proceedings of Machine Learning Research},
  volume = {235},
  pages = {13481--13544},
  year = {2024},
  publisher = {PMLR},
  url = {https://proceedings.mlr.press/v235/fernando24a.html}
}

@inproceedings{khattab2024dspy,
  title = {{DSPy}: Compiling Declarative Language Model Calls into State-of-the-Art Pipelines},
  author = {Khattab, Omar and Singhvi, Arnav and Maheshwari, Paridhi and Zhang, Zhiyuan and Santhanam, Keshav and Vardhamanan A, Sri and Haq, Saiful and Sharma, Ashutosh and Joshi, Thomas and Moazam, Hanna and Miller, Heather and Zaharia, Matei and Potts, Christopher},
  booktitle = {The Twelfth International Conference on Learning Representations},
  year = {2024},
  url = {https://proceedings.iclr.cc/paper_files/paper/2024/hash/f1cf02ce09757f57c3b93c0db83181e0-Abstract-Conference.html}
}

@inproceedings{li2023camel,
  title = {{CAMEL}: Communicative Agents for ``Mind'' Exploration of Large Language Model Society},
  author = {Li, Guohao and Hammoud, Hasan and Itani, Hani and Khizbullin, Dmitrii and Ghanem, Bernard},
  booktitle = {Advances in Neural Information Processing Systems},
  volume = {36},
  year = {2023},
  url = {https://proceedings.neurips.cc/paper/2023/hash/a3621ee907def47c1b952ade25c67698-Abstract-Conference.html}
}

@inproceedings{wu2024autogen,
  title = {{AutoGen}: Enabling Next-Gen {LLM} Applications via Multi-Agent Conversation},
  author = {Wu, Qingyun and Bansal, Gagan and Zhang, Jieyu and Wu, Yiran and Li, Beibin and Zhu, Erkang and Jiang, Li and Zhang, Xiaoyun and Zhang, Shaokun and Awadallah, Ahmed and White, Ryen W. and Burger, Doug and Wang, Chi},
  booktitle = {First Conference on Language Modeling},
  year = {2024},
  url = {https://arxiv.org/abs/2308.08155}
}

@inproceedings{qian2024chatdev,
  title = {{ChatDev}: Communicative Agents for Software Development},
  author = {Qian, Chen and Liu, Wei and Liu, Hongzhang and Chen, Nuo and Dang, Yufan and Li, Jiahao and Yang, Cheng and Chen, Weize and Su, Yusheng and Cong, Xin and Xu, Juyuan and Li, Dahai and Liu, Zhiyuan and Sun, Maosong},
  booktitle = {Proceedings of the 62nd Annual Meeting of the Association for Computational Linguistics (Volume 1: Long Papers)},
  pages = {15174--15186},
  year = {2024},
  publisher = {Association for Computational Linguistics},
  doi = {10.18653/v1/2024.acl-long.810},
  url = {https://aclanthology.org/2024.acl-long.810/}
}

@inproceedings{hong2024metagpt,
  title = {{MetaGPT}: Meta Programming for a Multi-Agent Collaborative Framework},
  author = {Hong, Sirui and Zhuge, Mingchen and Chen, Jonathan and Zheng, Xiawu and Cheng, Yuheng and Wang, Jinlin and Zhang, Ceyao and Wang, Zili and Yau, Steven and Lin, Zijuan and Zhou, Liyang and Ran, Chenyu and Xiao, Lingfeng and Wu, Chenglin and Schmidhuber, J{\"u}rgen},
  booktitle = {The Twelfth International Conference on Learning Representations},
  year = {2024},
  url = {https://proceedings.iclr.cc/paper_files/paper/2024/hash/6507b115562bb0a305f1958ccc87355a-Abstract-Conference.html}
}

@article{chen2023frugalgpt,
  title = {{FrugalGPT}: How to Use Large Language Models While Reducing Cost and Improving Performance},
  author = {Chen, Lingjiao and Zaharia, Matei and Zou, James},
  journal = {arXiv preprint arXiv:2305.05176},
  year = {2023},
  url = {https://arxiv.org/abs/2305.05176}
}

@inproceedings{zhuge2024gptswarm,
  title = {{GPTSwarm}: Language Agents as Optimizable Graphs},
  author = {Zhuge, Mingchen and Wang, Wenyi and Kirsch, Louis and Faccio, Francesco and Khizbullin, Dmitrii and Schmidhuber, J{\"u}rgen},
  booktitle = {Proceedings of the 41st International Conference on Machine Learning},
  series = {Proceedings of Machine Learning Research},
  volume = {235},
  pages = {62743--62767},
  year = {2024},
  publisher = {PMLR},
  url = {https://proceedings.mlr.press/v235/zhuge24a.html}
}

@inproceedings{zhang2025agentprune,
  title = {Cut the Crap: An Economical Communication Pipeline for {LLM}-based Multi-Agent Systems},
  author = {Zhang, Guibin and Yue, Yanwei and Li, Zhixun and Yun, Sukwon and Wan, Guancheng and Wang, Kun and Cheng, Dawei and Yu, Jeffrey and Chen, Tianlong},
  booktitle = {The Thirteenth International Conference on Learning Representations},
  year = {2025},
  url = {https://proceedings.iclr.cc/paper_files/paper/2025/hash/bbc461518c59a2a8d64e70e2c38c4a0e-Abstract-Conference.html}
}

@misc{openai2026gpt54,
  title = {{GPT-5.4 Model}},
  author = {{OpenAI}},
  year = {2026},
  url = {https://developers.openai.com/api/docs/models/gpt-5.4}
}

@misc{openai2026gpt54mini,
  title = {{GPT-5.4 Mini Model}},
  author = {{OpenAI}},
  year = {2026},
  url = {https://developers.openai.com/api/docs/models/gpt-5.4-mini}
}

@inproceedings{wang2025awm,
  title = {Agent Workflow Memory},
  author = {Wang, Zora Zhiruo and Mao, Jiayuan and Fried, Daniel and Neubig, Graham},
  booktitle = {Proceedings of the 42nd International Conference on Machine Learning},
  series = {Proceedings of Machine Learning Research},
  volume = {267},
  pages = {63897--63911},
  publisher = {PMLR},
  year = {2025},
  url = {https://proceedings.mlr.press/v267/wang25bx.html}
}

@article{xia2026skillrl,
  title = {{SkillRL}: Evolving Agents via Recursive Skill-Augmented Reinforcement Learning},
  author = {Xia, Peng and Chen, Jianwen and Wang, Hanyang and Liu, Jiaqi and Zeng, Kaide and Wang, Yu and Han, Siwei and Zhou, Yiyang and Zhao, Xujiang and Chen, Haifeng and Zheng, Zeyu and Xie, Cihang and Yao, Huaxiu},
  journal = {arXiv preprint arXiv:2602.08234},
  year = {2026},
  url = {https://arxiv.org/abs/2602.08234}
}

@article{ma2026skillgen,
  title = {{SkillGen}: Verified Inference-Time Agent Skill Synthesis},
  author = {Ma, Yuchen and Huang, Yue and Bao, Han and Zhuang, Haomin and Shukla, Swadheen and Galley, Michel and Zhang, Xiangliang and Feuerriegel, Stefan},
  journal = {arXiv preprint arXiv:2605.10999},
  year = {2026},
  url = {https://arxiv.org/abs/2605.10999}
}

@article{he2026skillcommit,
  title = {{SkillCommit}: Evolving Agent Skills through Behaviorally Validated Scope Expansion},
  author = {He, Yu and Yang, Weikai},
  journal = {arXiv preprint arXiv:2608.15165},
  year = {2026},
  url = {https://arxiv.org/abs/2608.15165}
}

@inproceedings{wang2025agentdropout,
  title = {{AgentDropout}: Dynamic Agent Elimination for Token-Efficient and High-Performance {LLM}-Based Multi-Agent Collaboration},
  author = {Wang, Zhexuan and Wang, Yutong and Liu, Xuebo and Ding, Liang and Zhang, Miao and Liu, Jie and Zhang, Min},
  booktitle = {Proceedings of the 63rd Annual Meeting of the Association for Computational Linguistics (Volume 1: Long Papers)},
  pages = {24013--24035},
  publisher = {Association for Computational Linguistics},
  year = {2025},
  doi = {10.18653/v1/2025.acl-long.1170},
  url = {https://aclanthology.org/2025.acl-long.1170/}
}

@inproceedings{chen2025optima,
  title = {Optima: Optimizing Effectiveness and Efficiency for {LLM}-Based Multi-Agent System},
  author = {Chen, Weize and Yuan, Jiarui and Qian, Chen and Yang, Cheng and Liu, Zhiyuan and Sun, Maosong},
  booktitle = {Findings of the Association for Computational Linguistics: ACL 2025},
  pages = {11534--11557},
  publisher = {Association for Computational Linguistics},
  year = {2025},
  doi = {10.18653/v1/2025.findings-acl.601},
  url = {https://aclanthology.org/2025.findings-acl.601/}
}
\bibliographystyle{plainnat}

\appendix

\section{Source Skill Evolution Protocol}
\label{app:source-evolution}

Table~\ref{tab:source-evolution-full} reports the source settings used by
transfer. Each evolved policy is compared with all-large
under the same model ladder, agent count, and source test instances. Prices
are test deployment costs. Source exploration is accounted for separately.
Values are rounded for display.

\begin{table}[ht]
\centering
\caption{Source test outcomes underlying transfer. CF quality is mean RMSE
(lower is better). AgentsNet quality is whole-network success/partial
correctness (higher is better). Prices are total test deployment USD and
exclude evolution. G and Q denote the GPT and Qwen ladders. $n=32$
AgentsNet uses generated graphs. The goal column is the recorded final
noninferiority, heterogeneous-tier, price, and confirmation gate.}
\label{tab:source-evolution-full}
\small
\begin{tabular*}{\linewidth}{@{\extracolsep{\fill}}llrrrrc@{}}
\toprule
Domain & Setting & Large Q & Evolved Q & Large \$ & Evolved \$ & Goal \\
\midrule
CF & G4  & 0.499 & 0.477 & 0.303 & 0.229 & yes \\
CF & G8  & 0.902 & 0.899 & 0.538 & 0.490 & yes \\
CF & G16 & 2.203 & 1.525 & 1.006 & 0.933 & yes \\
CF & G32 & 4.236 & 4.096 & 1.947 & 1.925 & yes \\
CF & Q4  & 0.667 & 0.568 & 0.227 & 0.171 & yes \\
CF & Q8  & 1.277 & 1.232 & 0.406 & 0.390 & yes \\
CF & Q16 & 3.518 & 1.809 & 0.762 & 0.714 & yes \\
CF & Q32 & 11.951 & 8.987 & 1.481 & 0.295 & yes \\
\midrule
AgentsNet & G4  & 0.800/0.928 & 0.800/0.928 & 0.522 & 0.458 & yes \\
AgentsNet & G8  & 0.467/0.875 & 0.467/0.875 & 2.231 & 2.053 & yes \\
AgentsNet & G16 & 0.533/0.870 & 0.467/0.860 & 8.549 & 8.254 & no \\
AgentsNet & G32 & 0.667/0.880 & 0.667/0.857 & 21.811 & 21.305 & yes \\
AgentsNet & Q4  & 0.867/0.949 & 0.933/0.960 & 0.418 & 0.385 & yes \\
AgentsNet & Q8  & 0.667/0.934 & 0.667/0.934 & 1.458 & 1.417 & yes \\
AgentsNet & Q16 & 0.467/0.907 & 0.400/0.840 & 4.452 & 4.277 & no \\
AgentsNet & Q32 & 0.200/0.638 & 0.200/0.638 & 17.475 & 16.792 & yes \\
\bottomrule
\end{tabular*}
\end{table}

The recorded source pools contain 20--153 skills, while the mapped candidate
pools contain 11--143 before a selector applies its quota
(Table~\ref{tab:skill-pool-counts}). The table separates routing from
communication skills so that the candidate count is not mistaken for the
number ultimately deployed.

\begin{table}[H]
\centering
\caption{Skill-pool size in the 16 source settings used for transfer. Source
counts include every recorded routing and communication skill, regardless of
status. Mapped counts are the candidates retained after transfer mapping,
before the selector quota is applied. R/C gives routing/communication counts.
Totals sum the two types. These are pool sizes, not deployed skill counts.}
\label{tab:skill-pool-counts}
\small
\begin{tabular*}{\linewidth}{@{\extracolsep{\fill}}llrrrr@{}}
\toprule
Domain & Setting & Source R/C & Source total & Mapped R/C & Mapped total \\
\midrule
CF & G4  & 15/7 & 22 & 6/7 & 13 \\
CF & G8  & 15/9 & 24 & 5/8 & 13 \\
CF & G16 & 15/6 & 21 & 5/6 & 11 \\
CF & G32 & 15/6 & 21 & 5/6 & 11 \\
CF & Q4  & 15/5 & 20 & 6/5 & 11 \\
CF & Q8  & 15/9 & 24 & 5/9 & 14 \\
CF & Q16 & 15/6 & 21 & 5/6 & 11 \\
CF & Q32 & 15/6 & 21 & 5/6 & 11 \\
\midrule
AgentsNet & G4  & 91/5 & 96 & 76/1 & 77 \\
AgentsNet & G8  & 148/5 & 153 & 135/5 & 140 \\
AgentsNet & G16 & 94/5 & 99 & 75/4 & 79 \\
AgentsNet & G32 & 73/4 & 77 & 51/2 & 53 \\
AgentsNet & Q4  & 91/5 & 96 & 74/0 & 74 \\
AgentsNet & Q8  & 148/5 & 153 & 139/4 & 143 \\
AgentsNet & Q16 & 94/5 & 99 & 74/5 & 79 \\
AgentsNet & Q32 & 73/4 & 77 & 49/4 & 53 \\
\bottomrule
\end{tabular*}
\end{table}

Each setting uses five exploration rounds. CF has 16 exploration, eight
search-validation, eight confirmation, and 32 test tasks. AgentsNet has 15
exploration, 15 validation, and 15 test tasks.

\section{Full Experimental Matrix}
\label{app:full-matrix}

Tables~\ref{tab:cf-rmse}--\ref{tab:agentsnet-evaluation-price-usd} report
all directions and the six frozen selectors, fixed parameter scaling,
source-deployed frozen bank, all-large execution, and Evo2Team.
DoRA-style is available in 20 cells.
The early four are separate runs, and the later sixteen are integrated in
the frozen suites. CF costs cover 32 test tasks. AgentsNet costs cover 15.
Saved tests of policies not selected at confirmation are marked as
diagnostics. Their held-out outcome labels do not make them accepted
deployments. Model-call estimates are not provider invoices.

\begin{table}[htbp]
\centering\small
\setlength{\tabcolsep}{2.4pt}
\caption{CF: Mean task RMSE (lower is better). G and Q denote GPT and Qwen ladders. -- denotes unavailable results.}
\label{tab:cf-rmse}
\begin{tabular}{lrrrrrrrrrr}
\toprule
Direction & Direct & KNN & KMM & CORAL & Options & SF-GPI & DoRA & Frozen & Large & Evo2Team \\
\midrule
$G4\!\to\!G8$ & 0.838 & 1.535 & 0.838 & 1.535 & 1.110 & 0.838 & 1.041 & 1.201 & 0.904 & 0.832 \\
$G8\!\to\!G16$ & 1.558 & 1.342 & 1.558 & 1.342 & 1.558 & 1.558 & 1.558 & 1.683 & 2.232 & 1.390 \\
$G16\!\to\!G32$ & 2.854 & 2.854 & 2.854 & 2.854 & 2.854 & 2.854 & 3.647 & 2.960 & 4.217 & 2.161 \\
$Q4\!\to\!Q8$ & 1.058 & 1.081 & 1.058 & 1.081 & 1.085 & 1.081 & 0.967 & 1.007 & 1.246 & 0.964 \\
$Q8\!\to\!Q16$ & 1.884 & 1.856 & 1.884 & 1.856 & 1.884 & 1.884 & 2.294 & 3.705 & 3.705 & 2.562 \\
$Q16\!\to\!Q32$ & 5.570 & 5.570 & 5.570 & 5.570 & 5.570 & 5.570 & 7.059 & 5.570 & 11.452 & 5.344 \\
$G4\!\to\!Q4$ & 0.625 & 1.081 & 0.625 & 1.081 & 0.625 & 0.634 & -- & 0.638 & 0.646 & 0.630 \\
$Q4\!\to\!G4$ & 0.478 & 0.478 & 0.478 & 0.478 & 0.554 & 0.478 & -- & 0.554 & 0.567 & 0.543 \\
$G8\!\to\!Q8$ & 1.003 & 1.804 & 1.003 & 1.804 & 0.972 & 1.003 & -- & 0.980 & 1.282 & 0.874 \\
$Q8\!\to\!G8$ & 0.743 & 1.557 & 0.743 & 1.557 & 0.743 & 0.743 & -- & 0.857 & 0.857 & 0.756 \\
$G16\!\to\!Q16$ & 1.858 & 1.858 & 1.858 & 1.858 & 1.858 & 1.858 & 4.267 & 1.858 & 3.476 & 1.741 \\
$Q16\!\to\!G16$ & 1.383 & 1.383 & 1.383 & 1.383 & 1.383 & 1.383 & 1.523 & 1.383 & 2.096 & 1.341 \\
$G32\!\to\!Q32$ & 3.377 & 3.377 & 3.377 & 3.377 & 3.377 & 3.377 & 3.265 & 11.713 & 11.713 & 3.440 \\
$Q32\!\to\!G32$ & 4.861 & 4.861 & 4.861 & 4.861 & 4.861 & 4.861 & 4.907 & 4.861 & 4.321 & 4.747 \\
\bottomrule
\end{tabular}
\end{table}

\begin{table}[htbp]
\centering\small
\setlength{\tabcolsep}{2.4pt}
\caption{CF: Total test deployment cost in USD. G and Q denote GPT and Qwen ladders. -- denotes unavailable results.}
\label{tab:cf-evaluation-price-usd}
\begin{tabular}{lrrrrrrrrrr}
\toprule
Direction & Direct & KNN & KMM & CORAL & Options & SF-GPI & DoRA & Frozen & Large & Evo2Team \\
\midrule
$G4\!\to\!G8$ & 0.754 & 0.306 & 0.754 & 0.306 & 0.681 & 0.804 & 0.642 & 0.449 & 0.538 & 1.005 \\
$G8\!\to\!G16$ & 2.124 & 1.557 & 2.124 & 1.557 & 2.131 & 2.131 & 2.279 & 0.962 & 1.005 & 2.127 \\
$G16\!\to\!G32$ & 2.763 & 2.763 & 2.763 & 2.763 & 2.763 & 2.763 & 2.955 & 1.852 & 1.947 & 1.577 \\
$Q4\!\to\!Q8$ & 0.432 & 0.576 & 0.432 & 0.576 & 0.731 & 0.576 & 0.434 & 0.345 & 0.406 & 0.335 \\
$Q8\!\to\!Q16$ & 1.639 & 0.816 & 1.639 & 0.816 & 1.658 & 1.658 & 1.707 & 0.746 & 0.762 & 1.162 \\
$Q16\!\to\!Q32$ & 1.381 & 1.381 & 1.381 & 1.381 & 1.381 & 1.381 & 1.429 & 1.381 & 1.481 & 1.381 \\
$G4\!\to\!Q4$ & 0.199 & 0.097 & 0.199 & 0.097 & 0.208 & 0.307 & -- & 0.157 & 0.227 & 0.400 \\
$Q4\!\to\!G4$ & 0.341 & 0.461 & 0.341 & 0.461 & 0.500 & 0.483 & -- & 0.259 & 0.302 & 0.259 \\
$G8\!\to\!Q8$ & 0.590 & 0.289 & 0.590 & 0.289 & 0.420 & 0.608 & -- & 0.328 & 0.406 & 0.694 \\
$Q8\!\to\!G8$ & 0.681 & 0.222 & 0.681 & 0.222 & 0.688 & 0.705 & -- & 0.521 & 0.538 & 1.305 \\
$G16\!\to\!Q16$ & 1.014 & 1.014 & 1.014 & 1.014 & 1.014 & 1.014 & 1.027 & 0.634 & 0.762 & 1.012 \\
$Q16\!\to\!G16$ & 1.088 & 1.088 & 1.088 & 1.088 & 1.088 & 1.088 & 1.119 & 1.088 & 1.005 & 1.069 \\
$G32\!\to\!Q32$ & 3.266 & 3.183 & 3.183 & 3.183 & 3.183 & 3.183 & 3.224 & 1.463 & 1.481 & 2.080 \\
$Q32\!\to\!G32$ & 0.715 & 0.715 & 0.715 & 0.715 & 0.715 & 0.715 & 0.731 & 0.715 & 1.990 & 0.711 \\
\bottomrule
\end{tabular}
\end{table}

\begin{table}[htbp]
\centering\small
\setlength{\tabcolsep}{2.4pt}
\caption{AGENTSNET: Whole-network success rate. G and Q denote GPT and Qwen ladders. -- denotes unavailable results.}
\label{tab:agentsnet-S-success-rate}
\begin{tabular}{lrrrrrrrrrr}
\toprule
Direction & Direct & KNN & KMM & CORAL & Options & SF-GPI & DoRA & Frozen & Large & Evo2Team \\
\midrule
$G4\!\to\!G8$ & 0.600 & 0.600 & 0.200 & 0.600 & 0.600 & 0.600 & 0.467 & 0.600 & 0.600 & 0.600 \\
$G8\!\to\!G16$ & 0.067 & 0.067 & 0.000 & 0.067 & 0.067 & 0.067 & 0.067 & 0.733 & 0.733 & 0.733 \\
$G16\!\to\!G32$ & 0.400 & 0.600 & 0.067 & 0.533 & 0.667 & 0.667 & 0.400 & 0.667 & 0.733 & 0.600 \\
$Q4\!\to\!Q8$ & 0.267 & 0.667 & 0.267 & 0.667 & 0.667 & 0.667 & 0.267 & 0.667 & 0.667 & 0.667 \\
$Q8\!\to\!Q16$ & 0.400 & 0.267 & 0.200 & 0.333 & 0.267 & 0.267 & 0.400 & 0.333 & 0.333 & 0.333 \\
$Q16\!\to\!Q32$ & 0.267 & 0.200 & 0.267 & 0.267 & 0.200 & 0.200 & 0.267 & 0.333 & 0.267 & 0.267 \\
$G4\!\to\!Q4$ & 0.733 & 0.800 & 0.333 & 0.800 & 0.733 & 0.733 & -- & 0.867 & 0.867 & 0.867 \\
$Q4\!\to\!G4$ & 0.267 & 0.467 & 0.333 & 0.467 & 0.600 & 0.600 & -- & 0.600 & 0.533 & 0.533 \\
$G8\!\to\!Q8$ & 0.200 & 0.600 & 0.267 & 0.600 & 0.200 & 0.200 & -- & 0.600 & 0.600 & 0.600 \\
$Q8\!\to\!G8$ & 0.600 & 0.600 & 0.200 & 0.600 & 0.533 & 0.533 & -- & 0.667 & 0.667 & 0.600 \\
$G16\!\to\!Q16$ & 0.333 & 0.400 & 0.200 & 0.400 & 0.467 & 0.467 & 0.400 & 0.533 & 0.533 & 0.533 \\
$Q16\!\to\!G16$ & 0.000 & 0.400 & 0.000 & 0.400 & 0.200 & 0.200 & 0.000 & 0.533 & 0.533 & 0.467 \\
$G32\!\to\!Q32$ & 0.200 & 0.267 & 0.200 & 0.267 & 0.533 & 0.533 & 0.200 & 0.267 & 0.267 & 0.267 \\
$Q32\!\to\!G32$ & 0.000 & 0.533 & 0.067 & 0.533 & 0.667 & 0.667 & 0.000 & 0.733 & 0.800 & 0.733 \\
\bottomrule
\end{tabular}
\end{table}

\begin{table}[htbp]
\centering\small
\setlength{\tabcolsep}{2.4pt}
\caption{AGENTSNET: Partial correctness. G and Q denote GPT and Qwen ladders. -- denotes unavailable results.}
\label{tab:agentsnet-P-partial-correctness}
\begin{tabular}{lrrrrrrrrrr}
\toprule
Direction & Direct & KNN & KMM & CORAL & Options & SF-GPI & DoRA & Frozen & Large & Evo2Team \\
\midrule
$G4\!\to\!G8$ & 0.894 & 0.894 & 0.474 & 0.894 & 0.894 & 0.894 & 0.745 & 0.894 & 0.894 & 0.894 \\
$G8\!\to\!G16$ & 0.436 & 0.252 & 0.346 & 0.252 & 0.252 & 0.252 & 0.412 & 0.952 & 0.952 & 0.952 \\
$G16\!\to\!G32$ & 0.637 & 0.830 & 0.425 & 0.817 & 0.889 & 0.889 & 0.687 & 0.897 & 0.964 & 0.895 \\
$Q4\!\to\!Q8$ & 0.681 & 0.943 & 0.675 & 0.943 & 0.943 & 0.943 & 0.619 & 0.943 & 0.943 & 0.943 \\
$Q8\!\to\!Q16$ & 0.817 & 0.781 & 0.542 & 0.848 & 0.781 & 0.781 & 0.788 & 0.848 & 0.848 & 0.848 \\
$Q16\!\to\!Q32$ & 0.604 & 0.481 & 0.621 & 0.714 & 0.491 & 0.491 & 0.442 & 0.794 & 0.727 & 0.727 \\
$G4\!\to\!Q4$ & 0.783 & 0.800 & 0.503 & 0.800 & 0.733 & 0.733 & -- & 0.867 & 0.867 & 0.867 \\
$Q4\!\to\!G4$ & 0.453 & 0.682 & 0.520 & 0.682 & 0.782 & 0.782 & -- & 0.782 & 0.749 & 0.749 \\
$G8\!\to\!Q8$ & 0.572 & 0.897 & 0.611 & 0.897 & 0.561 & 0.561 & -- & 0.897 & 0.897 & 0.897 \\
$Q8\!\to\!G8$ & 0.855 & 0.852 & 0.449 & 0.852 & 0.822 & 0.822 & -- & 0.919 & 0.919 & 0.912 \\
$G16\!\to\!Q16$ & 0.627 & 0.791 & 0.569 & 0.791 & 0.864 & 0.864 & 0.707 & 0.861 & 0.924 & 0.924 \\
$Q16\!\to\!G16$ & 0.334 & 0.742 & 0.334 & 0.742 & 0.529 & 0.512 & 0.311 & 0.880 & 0.880 & 0.876 \\
$G32\!\to\!Q32$ & 0.553 & 0.694 & 0.576 & 0.694 & 0.894 & 0.894 & 0.505 & 0.708 & 0.716 & 0.716 \\
$Q32\!\to\!G32$ & 0.317 & 0.803 & 0.379 & 0.803 & 0.836 & 0.836 & 0.318 & 0.897 & 0.963 & 0.897 \\
\bottomrule
\end{tabular}
\end{table}

\begin{table}[htbp]
\centering\small
\setlength{\tabcolsep}{2.4pt}
\caption{AGENTSNET: Total test deployment cost in USD. G and Q denote GPT and Qwen ladders. -- denotes unavailable results.}
\label{tab:agentsnet-evaluation-price-usd}
\begin{tabular}{lrrrrrrrrrr}
\toprule
Direction & Direct & KNN & KMM & CORAL & Options & SF-GPI & DoRA & Frozen & Large & Evo2Team \\
\midrule
$G4\!\to\!G8$ & 2.249 & 2.249 & 0.498 & 2.249 & 2.249 & 2.249 & 1.680 & 2.249 & 2.249 & 2.249 \\
$G8\!\to\!G16$ & 1.339 & 1.780 & 1.324 & 1.780 & 1.780 & 1.780 & 1.302 & 5.212 & 5.212 & 5.727 \\
$G16\!\to\!G32$ & 12.740 & 16.598 & 4.449 & 15.241 & 16.504 & 16.504 & 11.515 & 15.685 & 18.491 & 18.422 \\
$Q4\!\to\!Q8$ & 0.185 & 1.436 & 0.183 & 1.436 & 1.436 & 1.436 & 0.176 & 1.436 & 1.436 & 1.247 \\
$Q8\!\to\!Q16$ & 4.460 & 4.306 & 0.581 & 4.610 & 4.118 & 4.118 & 3.696 & 4.610 & 4.610 & 3.936 \\
$Q16\!\to\!Q32$ & 1.872 & 9.624 & 1.922 & 15.031 & 10.951 & 10.951 & 1.626 & 17.648 & 18.139 & 16.289 \\
$G4\!\to\!Q4$ & 0.317 & 0.374 & 0.049 & 0.374 & 0.366 & 0.366 & -- & 0.411 & 0.434 & 0.378 \\
$Q4\!\to\!G4$ & 0.107 & 0.397 & 0.126 & 0.397 & 0.422 & 0.422 & -- & 0.498 & 0.472 & 0.443 \\
$G8\!\to\!Q8$ & 0.163 & 1.284 & 0.179 & 1.285 & 0.795 & 0.795 & -- & 1.348 & 1.389 & 1.206 \\
$Q8\!\to\!G8$ & 1.851 & 1.899 & 0.469 & 1.899 & 2.116 & 2.116 & -- & 2.064 & 2.210 & 2.124 \\
$G16\!\to\!Q16$ & 3.245 & 2.012 & 0.513 & 3.730 & 4.154 & 4.154 & 3.218 & 4.039 & 4.487 & 3.867 \\
$Q16\!\to\!G16$ & 1.197 & 4.252 & 1.266 & 4.439 & 2.039 & 2.036 & 0.922 & 6.246 & 6.324 & 6.898 \\
$G32\!\to\!Q32$ & 1.619 & 12.487 & 1.702 & 12.487 & 17.106 & 17.106 & 1.584 & 16.551 & 17.762 & 15.820 \\
$Q32\!\to\!G32$ & 2.887 & 15.884 & 3.851 & 15.849 & 18.841 & 18.841 & 2.887 & 20.690 & 21.898 & 19.621 \\
\bottomrule
\end{tabular}
\end{table}

\section{Execution and Ablation Details}
\label{app:ablations}

\begin{table}[t]
\centering
\caption{Evo2Team outcomes and exploration cost relative to evolving a new
bank on the target. Fresh includes adaptation and confirmation. The final
column also charges reused reference runs once.}
\label{tab:coverage}
\normalsize
\begin{tabular*}{\linewidth}{@{\extracolsep{\fill}}lrrrrr@{}}
\toprule
Domain & Positive & Negative & Unfinished & \shortstack{Fresh /\\evolve} & \shortstack{With anchors /\\evolve} \\
\midrule
CF & 13 & 1 & 0 & 0.090--0.916 & 0.117--0.988 \\
AgentsNet & 7 & 7 & 0 & 0.497--0.719 & 0.593--0.846 \\
\bottomrule
\end{tabular*}
\end{table}

\begin{table}[t]
\centering
\caption{Evo2Team coverage and changes relative to all-large in AgentsNet. $S$ and $P$ changes are in percentage points. Positive cost saving is cheaper. P/N denote positive and negative held-out outcomes.}
\label{tab:agentsnet-coverage}
\normalsize
\begin{tabular*}{\linewidth}{@{\extracolsep{\fill}}llrrrr@{}}
\toprule
Direction & Outcome & Transferred & $\Delta S$ & $\Delta P$ & Saving (\%) \\
\midrule
$G4\to G8$ & N & 0/15 & 0.00 & 0.00 & 0.00 \\
$G8\to G16$ & N & 4/15 & 0.00 & 0.00 & -9.88 \\
$G16\to G32$ & N & 8/15 & -13.33 & -6.88 & 0.37 \\
$Q4\to Q8$ & P & 3/15 & 0.00 & 0.00 & 13.13 \\
$Q8\to Q16$ & P & 4/15 & 0.00 & 0.00 & 14.61 \\
$Q16\to Q32$ & N & 3/15 & 0.00 & 0.00 & 10.20 \\
$G4\to Q4$ & P & 3/15 & 0.00 & 0.00 & 13.02 \\
$Q4\to G4$ & P & 4/15 & 0.00 & 0.00 & 6.10 \\
$G8\to Q8$ & P & 6/15 & 0.00 & 0.00 & 13.15 \\
$Q8\to G8$ & N & 7/15 & -6.67 & -0.69 & 3.89 \\
$G16\to Q16$ & P & 3/15 & 0.00 & 0.00 & 13.82 \\
$G32\to Q32$ & P & 3/15 & 0.00 & 0.00 & 10.93 \\
$Q16\to G16$ & N & 5/15 & -6.67 & -0.42 & -9.06 \\
$Q32\to G32$ & N & 5/15 & -6.67 & -6.67 & 10.40 \\
\bottomrule
\end{tabular*}
\end{table}

\begin{table}[t]
\centering
\caption{These task records used transferred recipes, passed all three confirmation executions, and failed on a new graph in the same stratum. Test columns show source-deployed frozen reference $F$ $\to$ recipe. Each row is one task record.}
\label{tab:branch-failures}
\normalsize
\begin{tabular*}{\linewidth}{@{\extracolsep{\fill}}llcrr@{}}
\toprule
Direction & Task (stratum) & Confirm & Test $S$ & Test $P$ \\
\midrule
$Q4\to G4$ & Consensus (BA) & 3/3 & $1\to0$ & $1\to0.000$ \\
$Q8\to G8$ & Coloring (BA) & 3/3 & $1\to0$ & $1\to0.833$ \\
$Q16\to G16$ & Matching (WS) & 3/3 & $1\to0$ & $1\to0.938$ \\
\bottomrule
\end{tabular*}
\end{table}

\begin{table}[t]
\centering
\caption{Archived graph attributes for the four task records using transferred recipes that fail
on the new test instance. Arrows run from the fixed confirmation graph to
the test graph. Graph indices refer to official AgentsNet instances within
each task-type/topology/size stratum. These paired descriptions do not
separate graph-instance effects from fresh model responses.}
\label{tab:failed-graph-attributes}
\normalsize
\begin{tabular*}{\linewidth}{@{\extracolsep{\fill}}llrrrrr@{}}
\toprule
Direction & Task (graph index) & $n$ & $|E|$ & Max degree & Diameter & Rounds \\
\midrule
$Q4\to G4$ & Consensus BA ($g0\to g2$) & 4 & $4\to4$ & $3\to3$ & $2\to2$ & $5\to5$ \\
$Q8\to G8$ & Coloring WS ($g1\to g0$) & 8 & $16\to16$ & $6\to6$ & $2\to3$ & $5\to5$ \\
$Q8\to G8$ & Coloring BA ($g2\to g1$) & 8 & $12\to12$ & $5\to7$ & $3\to2$ & $5\to5$ \\
$Q16\to G16$ & Matching WS ($g0\to g2$) & 16 & $32\to32$ & $6\to7$ & $3\to4$ & $6\to6$ \\
\bottomrule
\end{tabular*}
\end{table}

\begin{figure}[htbp]
\centering
\includegraphics[width=0.92\linewidth]{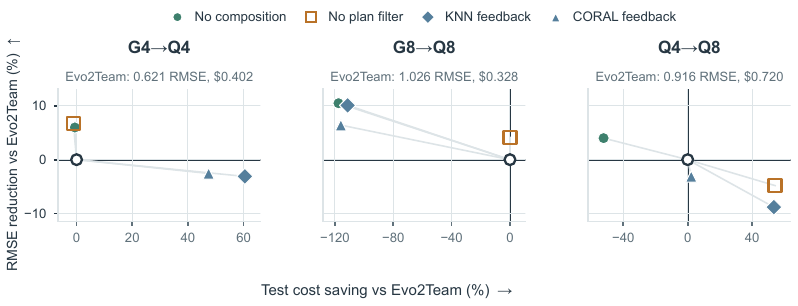}
\caption{\textbf{Qwen-target CF ablations.} Each panel compares four
interventions with Evo2Team (open origin) on 32 tasks. Rightward
means lower cost and upward lower RMSE; exact values are in
Table~\ref{tab:cf-ablation-complete}.}
\label{fig:ablation-qwen-cf}
\end{figure}

\begin{table}[H]
\centering\small
\caption{Complete Qwen-target CF ablations. Each variant has one recorded evaluation on the same 32 task IDs within its direction. Lower RMSE and lower test deployment cost are preferred.}
\label{tab:cf-ablation-complete}
\begin{tabular*}{\linewidth}{@{\extracolsep{\fill}}llrr@{}}
\toprule
Direction & Variant & RMSE & Cost (\$) \\
\midrule
$G4\to Q4$ & Evo2Team & 0.621 & 0.402 \\
 & No composition & 0.584 & 0.405 \\
 & No plan filter & 0.579 & 0.407 \\
 & KNN feedback & 0.640 & 0.159 \\
 & CORAL feedback & 0.637 & 0.211 \\
\midrule
$G8\to Q8$ & Evo2Team & 1.026 & 0.328 \\
 & No composition & 0.918 & 0.713 \\
 & No plan filter & 0.984 & 0.328 \\
 & KNN feedback & 0.922 & 0.693 \\
 & CORAL feedback & 0.960 & 0.708 \\
\midrule
$Q4\to Q8$ & Evo2Team & 0.916 & 0.720 \\
 & No composition & 0.879 & 1.097 \\
 & No plan filter & 0.960 & 0.329 \\
 & KNN feedback & 0.997 & 0.335 \\
 & CORAL feedback & 0.945 & 0.704 \\
\bottomrule
\end{tabular*}
\end{table}

\begingroup\small
\begin{longtable}{@{}lllrrr@{}}
\caption{Complete AgentsNet component ablations. Each variant has one recorded test evaluation on the same 15 graph tasks within its direction. $S$ is whole-team success, $P$ is mean partial correctness, and cost is test deployment USD.}\label{tab:agentsnet-ablation-complete}\\
\toprule
Target & Direction & Variant & $S$ & $P$ & Cost (\$) \\
\midrule
\endfirsthead
\toprule
Target & Direction & Variant & $S$ & $P$ & Cost (\$) \\
\midrule
\endhead
Qwen & $G4\to Q4$ & Evo2Team & 0.867 & 0.867 & 0.371 \\
 &  & No composition & 0.933 & 0.933 & 0.423 \\
 &  & No plan filter & 0.933 & 0.933 & 0.405 \\
 &  & No group guard & 0.867 & 0.867 & 0.434 \\
 &  & KNN feedback & 0.867 & 0.867 & 0.426 \\
 &  & CORAL feedback & 0.933 & 0.933 & 0.424 \\
\midrule
Qwen & $G8\to Q8$ & Evo2Team & 0.600 & 0.897 & 1.200 \\
 &  & No composition & 0.533 & 0.831 & 1.111 \\
 &  & No plan filter & 0.600 & 0.897 & 1.191 \\
 &  & No group guard & 0.600 & 0.897 & 1.389 \\
 &  & KNN feedback & 0.533 & 0.831 & 1.298 \\
 &  & CORAL feedback & 0.600 & 0.897 & 1.336 \\
\midrule
Qwen & $Q4\to Q8$ & Evo2Team & 0.667 & 0.943 & 1.246 \\
 &  & No composition & 0.600 & 0.935 & 1.455 \\
 &  & No plan filter & 0.733 & 0.949 & 1.233 \\
 &  & No group guard & 0.667 & 0.943 & 1.436 \\
 &  & KNN feedback & 0.667 & 0.943 & 1.436 \\
 &  & CORAL feedback & 0.667 & 0.943 & 1.420 \\
\midrule
GPT & $G4\to G8$ & Evo2Team & 0.467 & 0.761 & 1.687 \\
 &  & No composition & 0.600 & 0.894 & 2.249 \\
 &  & No plan filter & 0.600 & 0.894 & 2.249 \\
 &  & No group guard & 0.600 & 0.894 & 2.249 \\
 &  & KNN feedback & 0.600 & 0.894 & 2.249 \\
 &  & CORAL feedback & 0.467 & 0.698 & 1.552 \\
\midrule
GPT & $Q4\to G4$ & Evo2Team & 0.067 & 0.282 & 0.152 \\
 &  & No composition & 0.133 & 0.349 & 0.166 \\
 &  & No plan filter & 0.200 & 0.416 & 0.270 \\
 &  & No group guard & 0.533 & 0.749 & 0.472 \\
 &  & KNN feedback & 0.200 & 0.416 & 0.298 \\
 &  & CORAL feedback & 0.133 & 0.349 & 0.256 \\
\midrule
GPT & $Q8\to G8$ & Evo2Team & 0.333 & 0.585 & 1.195 \\
 &  & No composition & 0.400 & 0.652 & 1.392 \\
 &  & No plan filter & 0.267 & 0.519 & 1.095 \\
 &  & No group guard & 0.667 & 0.919 & 2.210 \\
 &  & KNN feedback & 0.667 & 0.919 & 2.210 \\
 &  & CORAL feedback & 0.267 & 0.460 & 0.934 \\
\bottomrule
\end{longtable}
\endgroup

The KNN-feedback and CORAL-feedback controls use target outcomes to search
subsets and are distinct from the frozen selectors. Evo2Team and both controls share
candidate-feedback ceilings of 80 adaptation plus 48 confirmation task
evaluations in CF, or 130 plus 90 in AgentsNet. Deduplication, fallback,
and confirmation can leave capacity unused, so actual API calls, tokens,
and exploration prices need not match.

\begin{figure}[htbp]
\centering
\includegraphics[width=\linewidth]{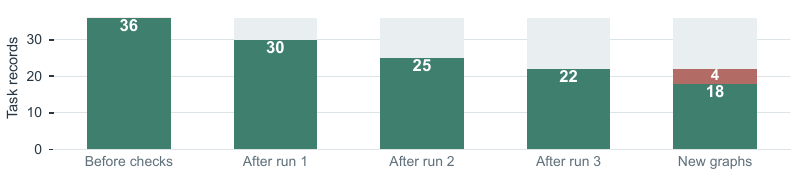}
\caption{\textbf{Confirmation attrition in five earlier AgentsNet directions.}
Of 36 task records initially using transferred recipes, 22 pass three
fixed-graph checks and 18 succeed on new graphs. Gray records were screened
out; red records fail in testing, which also resamples model responses.}
\label{fig:confirmation-outcomes}
\end{figure}

\begin{figure}[ht]
\centering
\includegraphics[width=0.70\linewidth]{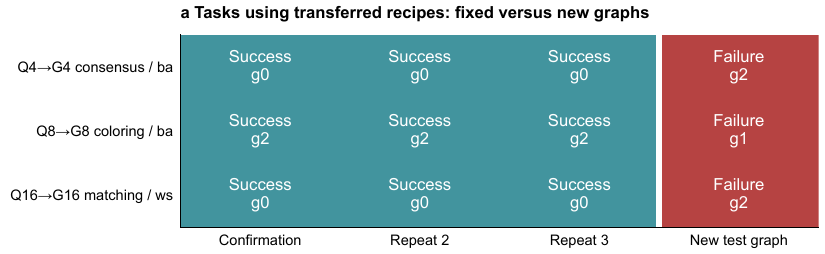}
\caption{\textbf{Confirmation-to-test trajectories.} Each row follows one
task record through three fixed-graph runs and one new-graph test;
Table~\ref{tab:branch-failures} gives partial-correctness changes.}
\label{fig:generalization}
\end{figure}

\begin{figure}[ht]
\centering
\includegraphics[width=0.70\linewidth]{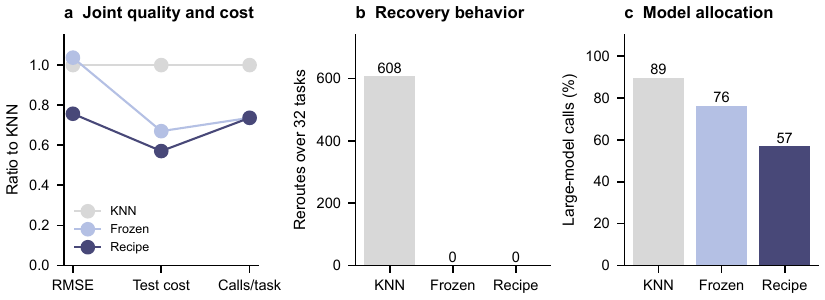}
\caption{\textbf{CF G16$\to$G32 execution.} KNN, the mapped source policy,
and Evo2Team share 32 test tasks. Panels compare quality and cost,
rerouting, and large-model call share; (a) uses KNN as denominator.}
\label{fig:execution-case}
\end{figure}

\begin{figure}[ht]
\centering
\includegraphics[width=0.70\linewidth]{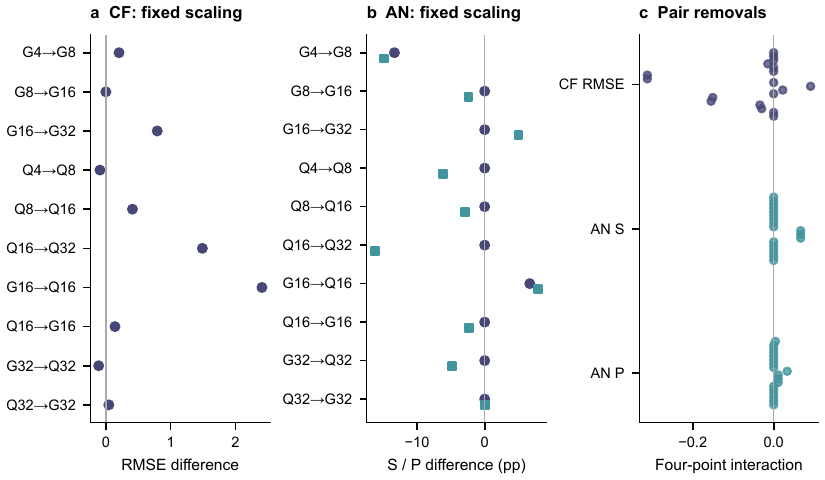}
\caption{\textbf{Parameter scaling and paired-skill probes.}
(a,b) DoRA-style minus Direct in ten cells per domain; lower RMSE and higher
$S/P$ are better. (c) Four-point interactions in 18 probes per domain.
These probes are separate from the full-run component ablations.}
\label{fig:ablations}
\end{figure}

The paired-probe analysis reconstructs all four evaluated recipes and checks
each difference against the stored interaction record. Probe selection and
parents depend on search history, and a parent or skill pair can recur.
The metric sign is not standardized into a universal synergy score: lower
RMSE and higher S/P are preferable, and the four-point difference is
conditional on the parent configuration and its observed responses.

The completed component ablations use separate full runs on matched task
sets. The paired-skill probes are not pooled into the main success counts.

\clearpage
\section{Confirmation Records}
\label{app:confirmation-records}

\paragraph{Guard transitions with recorded responses held fixed.}
For each of five accepted AgentsNet runs, we identify the selected candidate's
lineage through primary and repeated confirmation. Before each guard, active
task records are combined with the already static all-large records.
After the guard, only rejected groups are replaced by the recorded all-large
tasks. This is the implemented whole-task recomposition, requiring no
unobserved model response. We verify the rejected groups against the task-wise
S/P rule and the subsequent recipe ID. For G4$\to$Q4, the primary guard
raises S from 0.800 to 0.933 while cost changes from \$0.3735 to \$0.3796.
This quantifies protection purchased through fallback on the observed
confirmation tasks. It does not estimate what the complete search would
have produced without guards, or how rejected task groups would perform on test.

\begin{table}[htbp]
\centering\small
\caption{Trajectories for five earlier accepted AgentsNet directions. Columns count task records using transferred recipes before and after each guard. All 22 retained records had $S=1$ in three confirmation executions. Test counts include every retained record.}
\label{tab:guard-trajectory}
\begin{tabular}{lrrrrr}
\toprule
Direction & Initial & After 1 & After 2 & After 3 & Test successes \\
\midrule
$Q4\to Q8$ & 3 & 3 & 3 & 3 & 3/3 \\
$G4\to Q4$ & 6 & 4 & 3 & 3 & 3/3 \\
$Q4\to G4$ & 9 & 8 & 4 & 4 & 3/4 \\
$Q8\to G8$ & 9 & 8 & 8 & 7 & 5/7 \\
$Q16\to G16$ & 9 & 7 & 7 & 5 & 4/5 \\
\bottomrule
\end{tabular}
\end{table}

\paragraph{All retained task records, including successful ones.}
The five selected configurations start with 36 task records using transferred
recipes. Successive guards leave 30, 25, and 22. Among the final 22, every
record has $S=1$ on
the same confirmation graph in all three executions. New-graph successes
are 3/3, 3/3, 3/4, 5/7, and 4/5 in the directions listed in
Table~\ref{tab:guard-trajectory}. Q8$\to$G8 coloring WS accounts for the
fourth test failure, but both references also fail there. The three other
failures are the regressions in Table~\ref{tab:branch-failures}.
These are selected direction--task records with shared benchmark instances,
not independent Bernoulli trials. A comparison with confirmation on three
distinct graph instances cannot be recovered: each stratum provides one
adaptation, one confirmation, and one test instance in this protocol.
Reusing test graphs to tune such a comparator would change its evidence base.

\section{Cost Accounting and Deployment Amortization}
\label{app:cost-controls}

\paragraph{Charging static references.}
Let $A$ be fresh candidate adaptation and confirmation expenditure and
$H$ the logical cost of the reused adaptation and confirmation all-large
anchors. We report both $A$ and the sensitivity value $A+H$. The latter
charges each stored anchor once. Repeats reuse the same reference record,
not a fresh all-large execution. Across completed cells, $(A+H)/E$ remains
below one, where $E$ is recorded target-evolution expenditure. This
comparison uses existing settings and their reported price bases. It is not
a matched-token efficiency estimate or a provider-invoice reconciliation.
Source-pool construction remains outside
the per-transfer budget. Candidate calls and tokens are exported separately,
including all confirmation rounds, without imputing undocumented charges.

\begin{table}[htbp]
\centering\small
\caption{CF: sensitivity to charging reused static anchors once. Ratio is (fresh candidate search + anchors) / recorded target evolution. Break-even is in repeated test-sized batches versus all-large, reported only for positive-transfer cells with lower deployment cost. A dash does not represent zero.}
\label{tab:budget-cf}
\begin{tabular}{lrrrr}
\toprule
Direction & Fresh (\$) & Anchors (\$) & Charged ratio & Break-even \\
\midrule
$G4\to G8$ & 3.712 & 0.269 & 0.471 & -- \\
$G8\to G16$ & 6.664 & 0.504 & 0.396 & -- \\
$G16\to G32$ & 7.486 & 0.974 & 0.231 & 23 \\
$Q4\to Q8$ & 1.802 & 0.203 & 0.325 & 29 \\
$Q8\to Q16$ & 4.480 & 0.383 & 0.349 & -- \\
$Q16\to Q32$ & 5.279 & 0.741 & 0.587 & 60 \\
$G4\to Q4$ & 1.422 & 0.113 & 0.414 & -- \\
$Q4\to G4$ & 1.098 & 0.151 & 0.396 & 30 \\
$G8\to Q8$ & 1.799 & 0.203 & 0.325 & -- \\
$Q8\to G8$ & 4.063 & 0.269 & 0.512 & -- \\
$G16\to Q16$ & 3.345 & 0.382 & 0.267 & -- \\
$Q16\to G16$ & 3.970 & 0.504 & 0.247 & -- \\
$G32\to Q32$ & 9.389 & 0.741 & 0.988 & -- \\
$Q32\to G32$ & 3.307 & 0.974 & 0.117 & -- \\
\bottomrule
\end{tabular}
\end{table}

\begin{table}[htbp]
\centering\small
\caption{AGENTSNET: sensitivity to charging reused static anchors once. Ratio is (fresh candidate search + anchors) / recorded target evolution. Break-even is in repeated test-sized batches versus all-large, reported only for positive-transfer cells with lower deployment cost. A dash does not represent zero.}
\label{tab:budget-agentsnet}
\begin{tabular}{lrrrr}
\toprule
Direction & Fresh (\$) & Anchors (\$) & Charged ratio & Break-even \\
\midrule
$G4\to G8$ & 26.981 & 5.189 & 0.761 & -- \\
$G8\to G16$ & 83.065 & 15.030 & 0.787 & -- \\
$G16\to G32$ & 203.363 & 39.099 & 0.593 & -- \\
$Q4\to Q8$ & 14.811 & 3.325 & 0.735 & 97 \\
$Q8\to Q16$ & 51.035 & 9.035 & 0.814 & 90 \\
$Q16\to Q32$ & 168.216 & 35.485 & 0.771 & -- \\
$G4\to Q4$ & 3.395 & 0.830 & 0.711 & 75 \\
$Q4\to G4$ & 4.997 & 1.004 & 0.733 & 209 \\
$G8\to Q8$ & 16.766 & 3.210 & 0.809 & 110 \\
$Q8\to G8$ & 30.392 & 5.352 & 0.846 & -- \\
$G16\to Q16$ & 42.676 & 9.069 & 0.701 & 84 \\
$Q16\to G16$ & 86.684 & 15.047 & 0.816 & -- \\
$Q32\to G32$ & 207.771 & 36.445 & 0.597 & -- \\
\bottomrule
\end{tabular}
\end{table}

\paragraph{When lower deployment cost repays adaptation.}
For $M$ repetitions of the test-sized task mix, compare
$C_r(M)=A+H+M D_r$ with $C_b(M)=M D_b$, treating the baseline as already
available. If $D_b>D_r$, the first non-more-expensive batch is
$M^*=\lceil(A+H)/(D_b-D_r)\rceil$. Otherwise there is no monetary
break-even at the observed prices. These are arithmetic scenarios assuming
unchanged task mix, prices, quality, and no further adaptation, not measured
future savings. We report operational break-even only for accepted positive
transfers with lower deployment cost.

G4$\to$Q4, Q4$\to$Q8, and Q4$\to$G4 AgentsNet need respectively 75, 97,
and 209 test-sized batches against all-large. Against the source-deployed
bank, the same figures become 130, 97, and 111 batches. Evo2Team exploration
cost is lower than new target evolution in every completed direction, while
these break-even counts show when subsequent deployment savings recover it.
CF quality improvements can also be positive without lowering deployment
price.

\section{Implementation and Evaluation Details}
\label{app:implementation}

The following details retain the full transfer and evaluation protocol.
\subsection{Problem formulation}
\label{app-detail-sec:transfer-objects}

We represent a deployment setting by a model ladder $\mathcal{M}$ and an
agent count $n$. A ladder maps the relative tiers
$\{\mathrm{small},\mathrm{medium},\mathrm{large}\}$ to concrete models.
Given a source skill pool $\mathcal{S}_s$, Evo2Team constructs a target
configuration for $(\mathcal{M}_t,n_t)$. Each routing skill contains an
activation condition, a relative-tier action, and source evidence. Each
communication skill contains an action and its parameters, such as group
size, message budget, or recovery limits. Source evidence includes supporting
executions, counterexamples, and confidence. The pool is produced by source
exploration, then filtered for eligibility and redundant rules. Mapping
preserves relative-tier identities and updates agent-count applicability.

A recipe $r=(I,\theta)$ selects skill IDs $I\subseteq\mathcal{S}_s$ and
provides numeric parameter overrides $\theta$. Materialization applies these
overrides to copies of the selected source skills. The target planner $P_t$
then maps the recipe and task input $x$ to an execution plan, which the
executor $E_t$ realizes as a trace $\tau_r(x)$ and output $y_r(x)$:
\begin{equation}
  r=(I,\theta)
  \xrightarrow{\;P_t(\cdot,x)\;} p_r(x)
  \xrightarrow{\;E_t(\cdot,x)\;} (\tau_r(x),y_r(x)).
  \label{app-detail-eq:transfer-chain}
\end{equation}
The output is a frozen target recipe, or a context-conditioned dispatcher
over recipes. We use adaptation instances $\mathcal{D}_a$ to construct it,
confirmation instances $\mathcal{D}_c$ to select and guard it, and test
instances $\mathcal{D}_t$ to assess its generalization. The object of study is
the complete mapping from a selected configuration to its behavior and
outcomes, with the target model ladder held fixed when comparing methods.

\subsection{Stage I: From selected skills to execution plans}
\label{app-detail-sec:recipe-search}

Frozen transfer ranks the eligible source candidates, selects the top $K$,
and materializes them using the shared target planner. Rank determines
membership in the bank. It is not an execution weight. Target input features
are available to context-based selectors.

If $K$ is at least the eligible pool size, every ranking selects the same
bank. The materializer restores source-pool order and discards ranking
scores. When different banks are selected, activation conditions and action
precedence can still yield identical plans on particular inputs. We compare
each configuration's initial node-to-tier allocation and effective
communication parameters on adaptation tasks in a fixed order. Full node
traces then check realized execution, including decisions made after initial
planning. We compare these records by task and node, excluding method labels.
Identical initial plans do not determine later response-dependent decisions.

\subsection{Stage II: Adapting joint execution}
\label{app-detail-sec:composition}

\begin{figure}[t]
\centering
\includegraphics[width=0.70\linewidth]{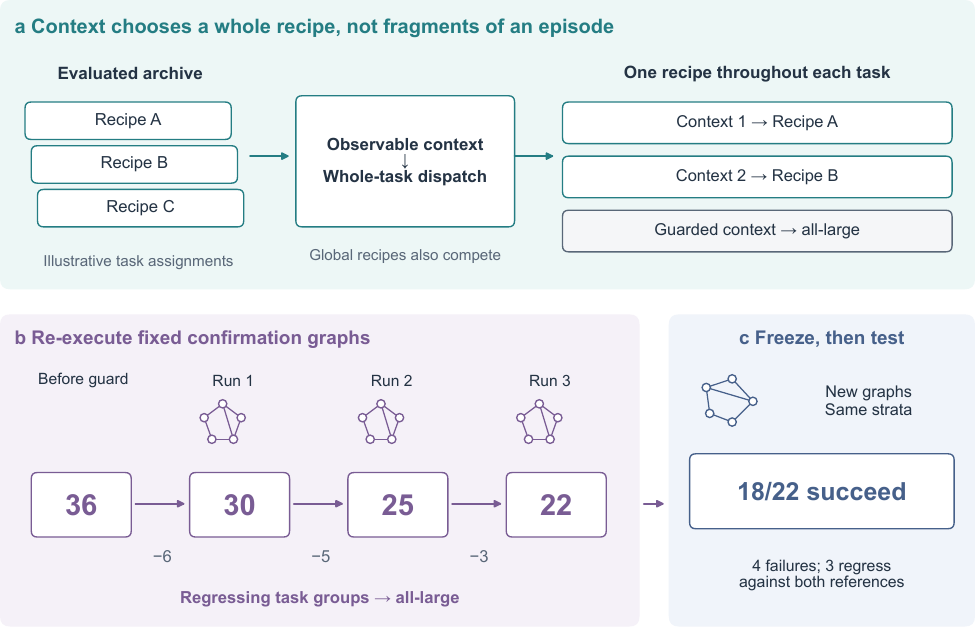}
\caption{\textbf{Whole-task composition and confirmation.}
(a) Context selects a complete recipe for each task. (b) Three fixed-graph
checks reduce transferred task records from 36 to 22; rejected groups use
all-large. (c) Of 22 retained records, 18 succeed on new graphs. Counts are
descriptive; Table~\ref{tab:guard-trajectory} gives each direction.}
\label{fig:context-guard-detail}
\end{figure}

\paragraph{Search over executable configurations.}
Evo2Team adapts the combinations and parameters through which
source skills affect the target workflow. Search starts from the complete
mapped source-deployed bank and a top-ranked source-candidate bank. Edits
add, remove, or replace skills, move routing interval boundaries, adjust
integer communication parameters, and recombine recipes. Structured proposals
remove competing communication slots or substitute routing rules with matching
conditions. Boundary proposals use midpoints between observed feature or
difficulty values to change which adaptation inputs activate a rule.

Before making target calls, search skips proposals that repeat an evaluated
initial plan on the adaptation tasks. Other proposals are screened on a subset of
$\mathcal{D}_a$, and the leading candidates advance to the full adaptation
set. Interleaving edit families allocates proposal slots across structural
and numeric changes. This procedure uses observable planning redundancy to
guide evaluation, while candidate quality is determined by actual execution.

\paragraph{Composition by task context.}
Different task groups can benefit from different recipes. We compose recipes
using observable context, choosing each task group's recipe from an archive of complete
adaptation episodes. A task executes one recipe throughout its workflow.
The composition score aggregates whole-task outcomes and costs. Keeping
episodes intact preserves dependencies among routing, communication, and
recovery decisions within each observed execution.

For Count-Frequency, the partition is either a single group or one balanced
split on predicted input difficulty, with at least three adaptation tasks
per group. For AgentsNet, selection first uses task-type groups, each
containing three topologies. A transferred task group is retained only if all
three adaptation instances of its task type succeed. The resulting policy
dispatches by task type and topology. Task groups failing the guard use the
all-large policy. The composed policy competes for the same confirmation
slots as global recipes, and confirmation can replace regressing task groups
with all-large execution.

\paragraph{Where the benefit occurs.}
To distinguish broad improvements from localized reuse, let $T$ be the set
of test instances assigned to transferred recipes and $N$ the total number
of instances. For a loss or cost $m_i$, define the per-task gain over the
all-large policy $L$ as $\Delta_i=m_i(L)-m_i(r)$. The coverage identity in
Eq.~\eqref{eq:coverage-gain} applies because tasks outside $T$ reuse their
all-large results and therefore have zero gain. We also retain individual
gains and losses, since a positive mean can combine improved and degraded
tasks.

\subsection{Stage III: Confirmation and held-out testing}
\label{app-detail-sec:confirmation}

Selection decides which configuration to deploy, while testing determines whether
its benefit extends to new instances. We prioritize quality gains, then
seek lower deployment cost among recipes with comparable quality. Let $F$
denote the source-deployed frozen bank evaluated on the target. We use the
quality-regression measure $g(r,b)$ from Eq.~\eqref{eq:regression}, where
$R$ is mean task RMSE, $S$ is whole-network success rate, and $P$ is mean
partial correctness. Thus $g<0$ indicates improvement in every quality
channel, and $g\leq\epsilon$ specifies the allowed regression.

The search ranks strict improvements over $L$ first, regardless of deployment
price. The second eligible group satisfies $g(r,L)\leq\epsilon$ and
$c(r)\leq(1+\eta)c(F)$, where $c$ is per-task deployment cost. Remaining
candidates are ranked by normalized quality and price violations, keeping
intermediate configurations available for further edits. Only the scalar
deployment price of $F$, obtained from the recorded target test reference,
enters selection. Its target quality is used after freezing.

\paragraph{Repeated confirmation and task-record retention.}
Shortlisted candidates are evaluated repeatedly on a separate confirmation
set $\mathcal{D}_c$. For Count-Frequency, the aggregate confirmation statistic
is $R_*(r)=\sqrt{J^{-1}\sum_{j=1}^{J}R_j(r)^2}$ across $J$ executions.
The quality-improvement route requires $R_*(r)<R(L)$, while the equivalent-quality
route allows a gap of $\epsilon$ and requires the price ceiling on adaptation
and every confirmation execution. Both routes bound each confirmation
regression by $2\epsilon$. AgentsNet applies the task-group guards across
confirmation executions and requires repeated quality improvement for the
performance-priority route. Confirmation revises which task records use a
transferred recipe before freezing the selected recipe and dispatcher for testing.

\paragraph{New-instance evaluation.}
\label{app-detail-sec:measurement}

Repeated confirmation holds the instance set fixed while resampling model
executions. Held-out testing changes the instances. We link each retained
task record's confirmation outcomes to its outcomes on $\mathcal{D}_t$, using the
same task type and topology for AgentsNet. This distinguishes execution
repeatability from performance on the recorded held-out instances. Paired task losses and
success transitions preserve task-level failures that an aggregate score can
offset with improvements elsewhere. Stage I identifies operational
differences, Stage II locates their benefits, and Stage III tests their reach.

We label completed Evo2Team outcomes after checking that adaptation costs
less than recorded target evolution. The held-out test offers three
routes: (i) strict quality improvement over both $L$ and $F$,
(ii) quality regression at most $\epsilon$ against both with
$c(r)\leq(1+\eta)c(F)$, or (iii) a cost-saving result with $c(r)<\min\{c(L),c(F)\}$ and similar quality to
both. For the last route, CF allows $R(r)-R(b)\leq\epsilon$ for each
$b\in\{L,F\}$. AgentsNet allows $S(b)-S(r)\leq1/15$, corresponding to
one task, and $P(b)-P(r)\leq\epsilon$ for each reference. The one-task
resolution adds Q4$\to$G4 to the positive set: its test success matches
all-large and trails $F$ by one task, while its deployment cost falls below
both. CF Q4$\to$G4 and AgentsNet G8$\to$Q8 likewise meet the held-out
quality and cost rule in saved tests, although their policies were not
selected at confirmation. These tests are marked as diagnostics, and the
remaining completed outcomes are negative. We compute the manuscript labels
from recorded test scores and prices using this rule.

\subsection{Extended experimental setup}
\label{app-detail-sec:setup}

The experiments address three linked questions. Do different selections
produce distinct recorded execution under a shared target executor? When
Evo2Team changes execution, how are its quality and cost gains
distributed across tasks? Do task records retained through repeated confirmation
succeed on new instances? We use the same transfer matrix to connect these
questions, tracking selections, execution records, and task outcomes rather
than treating each method's aggregate score as a complete account of transfer.

\subsection{Tasks and transfer matrix}
\label{app-detail-sec:tasks}

We evaluate two complementary settings: hierarchical aggregation in
Count-Frequency (CF) and decentralized coordination in AgentsNet
\citep{grotschla2025agentsnet}. CF asks agents to count integer frequencies
in disjoint shards and combine their outputs through merge and verification
stages. Each agent receives 64 elements sampled from 16 possible values, so
the total input grows as $64n$. Adaptation, confirmation, and test contain
8, 8, and 32 independently generated tasks, respectively.

AgentsNet includes coloring, matching, vertex cover, leader election, and
consensus on Watts--Strogatz, Barab\'asi--Albert, and Delaunay graphs. Each
node is an LLM agent, and messages follow existing edges and arrive in the next
synchronous round. We use three graph instances per topology, yielding
45 task--graph instances at each scale. One instance per
task-type/topology stratum goes to each of adaptation, confirmation, and test,
giving 15 tasks per split. Official graph instances are used at $n=4,8,16$.
$n=32$ extends the same task protocol with generated graphs.
Same-scale cross-ladder transfer includes paired source and target benchmark
instances. Held-out separation refers to target adaptation and confirmation.
Coloring, matching, and vertex cover use 4, 5, and 6 rounds at the three
official scales. Leader election and consensus use $2D+1$ rounds, where $D$
is graph diameter. This rule also applies to all five tasks at $n=32$.

The matrix covers bidirectional GPT--Qwen transfer at each
$n\in\{4,8,16,32\}$ and within-ladder scaling
$4\rightarrow8$, $8\rightarrow16$, and $16\rightarrow32$.
This gives 14 directions per task family and 28 domain--direction cells.
The study contains all 28 frozen-transfer cells and 28 completed
recipe-adaptation runs.

\subsection{Model ladders and source skills}
\label{app-detail-sec:models}

Table~\ref{app-detail-tab:model-ladders} lists the API model IDs stored in the
run configurations. In the GPT ladder, ``5.4'' identifies the GPT-5.4
release family and ``mini'' its smaller variant. The configured IDs
\texttt{gpt-5.4} and \texttt{gpt-5.4-mini} are aliases. OpenAI separately
lists date-pinned snapshots \texttt{gpt-5.4-2026-03-05} and
\texttt{gpt-5.4-mini-2026-03-17} \citep{openai2026gpt54,openai2026gpt54mini}.
The run configurations contain the aliases, so the dated snapshot used by
each request is not identified in these records. Relative tiers preserve
their labels during transfer, so
cross-ladder experiments replace the underlying models while retaining the
source rules' action vocabulary. Each cell starts from one source run's
routing and communication pools and its associated execution evidence.
Source pools are kept fixed during transfer. Target adaptation changes
their selection and numeric configuration. All language-model weights
remain fixed.

\begin{table}[t]
\centering
\caption{API model IDs recorded for the two experimental ladders. GPT IDs
without a date suffix are provider aliases, and the version number in each ID
names its release family.}
\label{app-detail-tab:model-ladders}
\small
\begin{tabular}{lll}
\toprule
Tier & GPT ladder & Qwen ladder \\
\midrule
Small & \texttt{gpt-4.1-nano} & \texttt{qwen3-8b} \\
Medium & \texttt{gpt-5.4-mini} & \texttt{qwen3.5-35b-a3b} \\
Large & \texttt{gpt-5.4} & \texttt{qwen3.7-max-2026-06-08} \\
\bottomrule
\end{tabular}
\end{table}

\subsection{Comparators and adaptation budget}
\label{app-detail-sec:baselines}

The frozen-selection suite contains Direct, KNN, KMM, CORAL, Options, and
SF--GPI. Direct ranks source reliability. KNN uses signed support from the
12 nearest source contexts. KMM weights source evidence to match target
contexts, while CORAL aligns source features before neighbor scoring. Options
scores activation coverage, reliability, and condition specificity. Our
one-step SF--GPI implementation scores source probe outcomes using success,
cost, latency, and failure penalties. These selectors use a common candidate pool
and executor. Early CF cells
use $K=6$, while later CF cells and all AgentsNet cells use $K=12$.

The DoRA-style comparator uses Direct's selected candidates and transforms
supported numeric fields $v$ according to
$v'=\Pi[b+\alpha(v-b)]$, with $\alpha=0.75$, fixed field-specific references
$b$, and projection $\Pi$ onto executable ranges, including integer rounding.
Actions and categorical fields are preserved. Its recorded coverage includes
the four early $4\rightarrow8$ domain/ladder cells and the 16 later
domain--direction cells. We additionally compare all-small, all-medium,
all-large, the default planner without transferred skills, and the mapped
source-deployed frozen bank $F$.

Evo2Team's adaptation uses two generations, at most six screened offspring per
generation, two promotions to full adaptation, and two confirmation slots.
Screening uses four CF tasks or five AgentsNet tasks, with the latter covering
all five task types. Candidate recipes receive three confirmation executions
on the same held-out confirmation instances. The assembly quota is 12. The
complete source-deployed starting bank is retained when larger. We fix
$\epsilon=0.05$ and $\eta=0.02$ across all cells.

\subsection{Metrics, accounting, and evaluation units}
\label{app-detail-sec:metrics}

CF reports the arithmetic mean of task-level RMSE, where each task's RMSE is
computed over the union of predicted and reference histogram bins. AgentsNet
reports whole-network binary success averaged over tasks ($S$) and the
benchmark's task-specific partial correctness averaged over tasks ($P$).
We pair methods by test task ID and report task wins, losses, and ties for
CF and success/partial-score transitions for AgentsNet. Execution diagnostics
include tier allocations, model calls, grouping, rerouting, and the number
of task records using transferred recipes.

API usage is separated into source evolution, target adaptation and
confirmation, static-reference acquisition, and test deployment. The
adaptation-cost comparison charges fresh candidate calls and reports reused
static all-large anchors separately. Candidate response caches are isolated
by recipe, stage, and repetition. Frozen baselines share a response cache
within a suite, making identical requests directly comparable. At test time,
tasks assigned to all-large can reuse recorded static results. Their logical
deployment cost remains included even when no new call is made.

The task instance is the unit of outcome comparison. Repeated confirmation
resamples executions of fixed instances. The test split measures transfer to
different instances within the same strata. Shared tasks across transfer
directions are tracked as repeated comparisons. The generated $n=32$ graphs
are reported as a scale extension.
Abstention is a selection outcome: saved test evaluations of rejected
recipes are retained as diagnostics and separated from accepted transfer.

\end{document}